\def\year{2022}\relax
\documentclass[letterpaper]{article} 

\usepackage{amsmath,amsfonts,bm}

\def\eqref#1{equation~\ref{#1}}

\def\1{\bm{1}}

\DeclareMathAlphabet{\mathsfit}{\encodingdefault}{\sfdefault}{m}{sl}
\SetMathAlphabet{\mathsfit}{bold}{\encodingdefault}{\sfdefault}{bx}{n}

\newcommand{\E}{\mathbb{E}}

\newcommand{\R}{\mathbb{R}}

\DeclareMathOperator*{\argmax}{arg\,max}

\usepackage{aaai23}  
\usepackage{times}  
\usepackage{helvet}  
\usepackage{courier}  
\usepackage[hyphens]{url}  
\usepackage{graphicx} 
\usepackage{natbib}  
\usepackage{caption} 
\DeclareCaptionStyle{ruled}{labelfont=normalfont,labelsep=colon,strut=off} 
\usepackage{algorithm}
\usepackage{algorithmic}
\usepackage{amsmath}
\usepackage{amssymb}
\usepackage{bbm}
\usepackage{multicol}
\usepackage{graphicx,wrapfig,lipsum}
\usepackage{amsthm}
\usepackage{stackengine}
\usepackage{subfig}
\usepackage{wrapfig}
\usepackage{booktabs}       

\theoremstyle{plain}
\newtheorem{theorem}{Theorem}[section]

\theoremstyle{definition}
\newtheorem{definition}[theorem]{Definition}

\theoremstyle{remark}

\usepackage{newfloat}
\usepackage{listings}
\floatstyle{ruled}
\newfloat{listing}{tb}{lst}{}
\floatname{listing}{Listing}
\nocopyright
\title{Adversarial Robust Deep Reinforcement Learning Requires Redefining Robustness}

\author{
Ezgi Korkmaz
}
\affiliations{
}

\usepackage{bibentry}

\begin{document}

\maketitle

\begin{abstract}
Learning from raw high dimensional data via interaction with a given environment has been effectively achieved through the utilization of deep neural networks. Yet the observed degradation in policy performance caused by imperceptible worst-case policy dependent translations along high sensitivity directions (i.e. adversarial perturbations) raises concerns on the robustness of deep reinforcement learning policies. In our paper, we show that these high sensitivity directions do not lie only along particular worst-case directions, but rather are more abundant in the deep neural policy landscape and can be found via more natural means in a black-box setting. Furthermore, we show that vanilla training techniques intriguingly result in learning more robust policies compared to the policies learnt via the state-of-the-art adversarial training techniques. We believe our work lays out intriguing properties of the deep reinforcement learning policy manifold and our results can help to build robust and generalizable deep reinforcement learning policies.
\end{abstract}

\section{Introduction}
\label{submission}

Following the initial work of \citet{mn15}, the use of deep neural networks as function approximators in reinforcement learning has led to a dramatic increase in the capabilities of reinforcement learning policies \citep{schulman17,vin19,julian20}. In particular, these developments allow for the direct learning of strong policies from raw, high-dimensional inputs (i.e. visual observations). With the successes of these new methods come new challenges regarding the robustness and generalization capabilities of deep reinforcement learning agents.

Initially, \citet{szegedy13} showed that specifically crafted \textit{imperceptible} perturbations can lead to misclassification in image classification. After this initial work a new research area emerged to investigate the abilities of deep neural networks against specifically crafted adversarial examples. While various works studied many different ways to compute these examples \citep{carlini17, madry18, fellow15, kurakin16}, several works focused on studying ways to increase the robustness against such specifically crafted perturbations, based on training with the existence of such perturbations \citep{madry18, florian18, fellow15, xie20}.

As image classification suffered from this vulnerability towards worst-case distributional shift in the input, a series of work conducted in deep reinforcement learning showed that deep neural policies are also susceptible to specifically crafted imperceptible perturbations \citep{huang17, kos17, pattanaik17, lin17,korkmaz20, sun20, korkmazuai}. While one line of work put effort on exploring these vulnerabilities in deep neural policies, another line in parallel focused making them robust and reliable via adversarial training \citep{pinto17, mandlekar17, glaeve19}.

While adversarial perturbations and adversarial training provide a notion of robustness for trained deep neural policies, in this paper we approach the resilience problem of deep reinforcement learning from a wider perspective, and propose to investigate the deep neural policy manifold along high-sensitivity directions. Along this line we essentially seek answers for the following questions:
\begin{itemize}
\item \textit{How can we probe the deep neural policy decision boundary with policy-independent high-sensitivity directions innate to the MDP within the perceptual similarity bound?}
\item \textit{Is it possible to affect the state-of-the-art deep reinforcement learning policy performance trained in high-dimensional state representation MDPs with policy-independent high-sensitivity directions intrinsic to the MDP?}
\item \textit{What are the effects of state-of-the-art certified adversarial training on the robustness of the policy compared to straightforward vanilla training when policy-independent high-sensitivity directions are present?}
\end{itemize}
Thus, to be able answer these questions, in this work we focus on the notion of robustness for deep reinforcement learning policies and make the following contributions:

\begin{itemize}
\item We probe the deep reinforcement learning manifold via policy dependent and policy-independent high-sensitivity directions innate to the MDP.
\item We run multiple experiments in the Arcade Learning Environment (ALE) in various games with high dimensional state representation and provide the relationship between the perceptual similarities to base states under policy dependent and policy-independent high-sensitivity directions.
\item We compare policy-independent high-sensitivity directions with the state-of-the-art adversarial directions based on $\ell_p$-norm changes, and we show that policy-independent high-sensitivity directions intrinsic to the MDP are competitive in degrading the performance of the deep reinforcement learning policies with lower perceptual similarity distance. Thus, the results of this contradistinction of adversarial directions and policy-independent high-sensitivity directions intrinsic to the MDP evidently demonstrates the abundance of high-sensitivity directions in the deep reinforcement learning policy manifold.
\item Finally, we inspect state-of-the-art adversarial training under changes intrinsic to the MDP, and demonstrate that the adversarially trained models become more vulnerable to several different types of policy-independent high-sensitivity directions compared to vanilla trained models.
\end{itemize}

\section{Background and Related Work}
\subsection{Preliminaries}
\label{prelim}
In this paper we consider Markov Decision Processes (MDPs) given by a tuple $ (S,A,\mathcal{T},r,\gamma,s_i)$. The reinforcement learning agent interacts with the MDP by observing states $s \in S$, taking actions $a \in A$ and receiving rewards $r(s,a,s')$. Here $s_i$ represents the initial state of the agent, and $\gamma \in (0,1]$ represents the discount factor. The probability of transitioning to state $s'$ when the agent takes action $a$ in state $s$ is determined by the Markovian transition kernel $\mathcal{T}: S \times A \times S \to \R$. The reward received by the agent when taking action $a$ in state $s$ is given by the reward function $r: S \times A \times S \to \R$. The goal of the agent is to learn a policy $\pi: S \times A \to \R$ which takes an action $a$ in state $s$ that maximizes the expected cumulative discounted reward $\sum_{t=0}^{T-1}\gamma^t r(s_t,a_t, s_{t+1})$ that the agent receives via interacting with the environment.
\begin{equation}
\tilde{\pi} = \argmax_{\pi} \sum_t \E_{s_t,a_t \sim \mathbb{P}_\pi} [r(s_t,a_t, s_{t+1})]
\end{equation}
where $\mathbb{P}_\pi$ represents the occupancy distribution of the trajectory followed by the policy $\pi(a_t|s_t)$. Hence, this goal can be achieved via learning the state-action value function via iterative Bellman update
\[
Q(s,a) = \E_\pi [\sum_{t=0}^{T-1}\gamma^t r(s_t,a_t, s_{t+1}) |s_i=s, a_i=a]
\]
assigning a value to each state-action pair. In high dimensional state representation MDPs the state-action values are estimated via function approximators.
\begin{align*}
\theta_{t+1} = \theta_t + \alpha (Q_t^{\textrm{target}} - Q(s_t,a_t; \theta_t)) \nabla_{\theta_t} Q(s_t,a_t; \theta_t)
\end{align*}
where $Q_t^{\textrm{target}}$ is $r(s_t,a_t,s_{t+1}) +\gamma \max_a Q(s_{t+1},a;\theta_t)$.

\subsection{Computing Adversarial Directions}
\label{advpert}

\citet{szegedy13} proposed to minimize the distance between the base image and adversarially produced image to create adversarial directions. The authors used box-constrained L-BFGS to solve this optimization problem. \citet{fellow15} introduced the fast gradient method (FGM),
\begin{equation}
\mathnormal{\displaystyle x_{\textrm{adv}} = x+ \epsilon \cdot \frac{\nabla_{x}J(\displaystyle x,y)}{||\nabla_{x}J(x,y)||_p},}
\end{equation}
for crafting adversarial examples in image classification by taking the gradient of the cost function $J(x,y)$ used to train the neural network in the direction of the input, where $x$ is the input, $y$ is the output label, and $J(x,y)$ is the cost function. \citet{carlini17} introduced targeted attacks in the image classification domain based on distance minimization between the adversarial image and the base image while targeting a particular label. Thus, in deep reinforcement learning the \citet{carlini17} formulation will find the minimum distance to a nearby state in an $\epsilon$-ball $\mathcal{D}_{\epsilon,p}(s)$ such that,
\begin{align*}
& \min_{\hat{s} \in \mathcal{D}_{\epsilon,p}(s)} \mathnormal{\|\hat{s}-s\|_p} \\
\text{subject to} \:\:&\mathnormal{\argmax_a Q(s,a) \neq \argmax_a Q(\hat{s},a) }
\label{lin}
\end{align*}
where $s \in S$ represents the base state, $\hat{s} \in \mathcal{D}_{\epsilon,p}(s)$ represents the state when it is moved along the adversarial directions.
This formulation attempts to minimize the distance to the base state, constrained to states leading to sub-optimal actions as determined by the $Q$-network. Note that the Carlini \& Wagner formulation has quite recently been used to demonstrate that the state-of-the-art adversarial trained policies share similar, and even in some cases identical, adversarial directions with the vanilla trained deep reinforcement learning policies \cite{korkmaz2022aaai}.
In contrast to adversarial attacks, in our proposed threat model we will not need any information on the cost function used to train the network, the $Q$-network of the trained agent, or access to the visited states themselves.

\subsection{Adversarial Approach in Deep Reinforcement Learning}
\label{advrl}

The first adversarial attacks on deep reinforcement learning introduced by \citet{huang17} and \citet{kos17} adapted FGSM from image classification to the deep reinforcement learning setting.
Subsequently, \citet{pinto17} and \citet{glaeve19} focused on modeling the interaction between the adversary and the agent as a zero-sum Markov game, while \citet{lin17,sun20} focused on strategically timing when (i.e. in which state) to attack an agent using perturbations computed with the Carlini \& Wagner adversarial formulation.
Orthogonal to this line of research some studies demonstrated that deep reinforcement learning policies learn adversarial directions from underlying MDPs that are shared across states, across MDPs and across algorithms \cite{korkmaz2022aaai}.
While proposing novel techniques to uncover non-robust features, some recent studies demonstrated the persistent existence of the non-robust features in state-of-the-art adversarial training methods\footnote{See \cite{korkmaz21cvpr} for inaccuracy and inconsistency of the state-action value function learnt by adversarially trained policies. For more on robustness problems in inverse deep reinforcement learning see \cite{korkmazinverserlicml,korkmazspectralneurips}} \cite{korkmazuai}.

\subsection{Perceptual Similarity Distance}
\label{perceptualsim}

Internal activations of networks trained for high-level tasks correspond to human perceptual judgements across different network architectures \cite{alex12,karen14,ian16} without calibration \cite{zhang18}. More importantly, it is possible to measure the perceptual similarity distance between two images with LPIPS matching human perception. Thus, in our experiments we measure the distance of moving along the high sensitivity directions from the base states with LPIPS. In particular, $\mathcal{P}_{\textrm{similarity}}(s,\hat{s})$ returns the distance between $s$ and $\hat{s}$ based on network activations, and results in an effective approximation of human perception.
In more detail, the LPIPS metric is given by measuring the $\ell_2$-distance between a normalized version of the activations of the neural network at several internal layers. For each layer $l$ let $W_l$ be the width, $H_l$ the height, and $C_l$ the number of channels. Further, let $y^l \in \R^{W_l\times H_l\times C_l}$ denote the vector of activations in convolutional layer $l$.
To compute the perceptual similarity distance between two states $s$ and $\hat{s}$, first calculate the channel-normalized internal activations $\hat{y}_s^l,\hat{y}_{\hat{s}}^l \in \R^{W_l\times H_l\times C_l}$ (corresponding to $s$ and $\hat{s}$ respectively) for $L$ internal layers, and scale each channel in $\hat{y}_s^l$ and $\hat{y}_{\hat{s}}^l$ by the same, fixed weight vector $w_l\in \R^{C_l}$.
The last step is then to compute the perceptual similarity distance by first averaging the $\ell_2$-distance between the scaled activations over the spatial dimensions, and then summing over the $L$ layers.

\section{Moving Through the Deep Neural Policy Manifold via High-Sensitivity Directions}

\label{main}
To investigate the deep neural policy manifold we will probe the deep reinforcement learning decision boundary via both adversarial directions and directions innate to the state representations. While the adversarial directions are specifically optimized high-sensitivity directions in the deep neural policy landscape (i.e. worst-case distributional shift) within an \textit{imperceptibility} bound as described in Section \ref{advrl}, the natural directions represents intrinsic semantic changes in the state representations within the imperceptibility distance.

\begin{definition}
	\label{def:highsensitivity}
	Let $\pi$ be a policy in an MDP $\mathcal{M}$ and let $S$ be the set of states in $\mathcal{M}$. Let $\epsilon, \delta >0$. An  $(\epsilon,\delta)$-high-sensitivity direction function for $\pi$ is a function $\xi(s,\pi)$ taking values in $S$ such that $\mathcal{P}_{\textrm{similarity}}(s, s+\xi(s,\pi)) \leq \epsilon$ for all $s\in S$, and
	\begin{align*}
		\mathbb{E}_{a_t \sim \pi(s_t + \xi(s_t,\pi),\cdot)} &\left[\sum_{t=0}^{T-1}\gamma^t r(s_t,  a_t, s_{t+1})\right] \\
		&< \delta \cdot \mathbb{E}_{a_t \sim \pi(s_t,\cdot)} \left[\sum_{t=0}^{T-1}\gamma^t r(s_t,a_t,s_{t+1})\right].
	\end{align*}
\end{definition}
Intuitively, $\xi(s,\pi)$ is a high-sensitivity direction function if translating by $\xi(s,\pi)$ in state $s$ causes a significant drop in expected cumulative rewards when executing the policy $\pi$. Note that the function $\xi(s,\pi)$ in Definition \ref{def:highsensitivity} takes the policy $\pi$ as input, and so is able to use information about the behavior of $\pi$ in state $s$ in order to compute the direction $\xi$.
We next introduce a restricted version of Definition \ref{def:highsensitivity} where the function is not allowed to use any information about $\pi$.

\begin{definition}
  \label{def:natural}
	Let $S$ be the set of states for an MDP $\mathcal{M}$, let $\pi \in \Pi$ be a set of policies in $\mathcal{M}$, and let $\xi:S \to S$ be a function on $S$.
	Let $\epsilon, \delta >0$.
	$\xi(s)$ is a fixed $(\epsilon,\delta)$-high-sensitivity direction function if the function $\phi(s,\pi) = \xi(s)$ is an $(\epsilon,\delta)$-high-sensitivity direction function for all $\pi \in \Pi$.
\end{definition}

\begin{algorithm}[t]
\caption{Probing Neural Manifold with High-sensitivity Directions within Perceptual Similarity}
\label{alg}
\begin{algorithmic}
\STATE {\bfseries Input:} Policy $\pi(s,a)$, high-sensitivity direction function $\xi(s,\pi)$, internal activations in convolutional layer $y^l \in \R^{W_l\times H_l\times C_l}$, parameter $\epsilon,\delta > 0$.
\FOR{$t=0$ {\bfseries to} $T$}
\STATE $a_t = \argmax_{a' \in A(s)} \pi(s_t + \xi(s_t,\pi), a')$
\STATE Sample $s_{t+1} \sim \mathcal{T}(s_t,a_t, \cdot)$
\STATE $\mathcal{P}_{\textrm{similarity}}(s,s+\xi(s,\pi)) = $
\STATE $ \qquad \sum_l\frac{1}{H_l W_l}\sum_{h,w} \lVert w_l \odot (\hat{y}^l_{shw} - \hat{y}^l_{(s+\xi(s,\pi))hw}) \rVert_2^2$
\STATE $\mathcal{PS} += \mathcal{P}_{\textrm{similarity}}(s_t,s_t + \xi(s_t,\pi))$
\STATE $\mathcal{R} += r(s_t,a_t)$
\ENDFOR
\STATE {\bfseries Return:} Total reward $\mathcal{R}$ and average perceptual similarity $\frac{\mathcal{PS}}{T}$.
\end{algorithmic}
\end{algorithm}

\begin{table*}[t!]
\caption{Impacts on the policy performance, perceptual similarity distances $\mathcal{P}_{\textrm{similarity}}$ to the base states, and raw scores for \citet{carlini17} formulation and policy-independent high-sensitivity directions innate to the environment. We report all of the results with the standard error of the mean.}
\vskip -0.1in
\label{all}
\centering
\scalebox{0.81}{
\begin{tabular}{lccccccr}
\toprule
ALE MDPs     & BankHeist
											      				 & JamesBond
											      				 & Pong
											      				 & Riverraid
											      				 & TimePilot \\
\midrule
C\&W Impact   &  0.982$\pm$0.009
													& 0.451$\pm$0.231
													& 0.995$\pm$0.014
													& 0.928$\pm$0.030
													&  0.567 $\pm$0.159 &  \\
B\&C Impact  &0.966$\pm$ 0.030
														& 0.913	$\pm$0.047
           											     &1.0$\pm$0.009
           											     & 0.951 $\pm$0.016
              											 & 0.663$\pm$0.239 \\
Blurred Observations Impact                       & 0.979$\pm$0.009
                 											& 0.635$\pm$0.200
                       								 &1.0$\pm$0.000
                       								 & 0.946$\pm$0.015
                       								 & 0.589$\pm$0.150 \\
Rotation Impact                      &  0.997$\pm$0.004
                                    &  0.635$\pm$0.189
                                    &  0.99$\pm$0.015
                                &0.942$\pm$0.042
                              & 0.581$\pm$0.158 \\
Shifting Impact                      &  0.985   $\pm$0.005
                                   	&  0.865$\pm$0.140
                                    & 1.0$\pm$0.00
                                    &   0.935 $\pm$0.023
                                   &  0.623$\pm$0.199     \\
DCT Artifacts Impact        & 0.980 $\pm$0.013
                                    &0.884 $\pm$0.128
                                    &  0.962$\pm$0.032
                                   & 0.803 $\pm$0.051
                                   & 0.578 $\pm$0.271    \\
PT Impact                  &   0.998$\pm$0.003
                                   						&  0.865$\pm$0.087
                                             &  0.996$\pm$0.009
                                            &  0.968$\pm$0.006
                                          &   0.624$\pm$0.198     \\
\midrule
C\&W $\mathcal{P}_{\textrm{similarity}}$ & 0.0657$\pm$0.0073
                                                      & 0.2622$\pm$0.0312
                                          				    & 0.6134$\pm$0.0271
                                                      & 0.2714$\pm$0.0285
                                                      & 0.1336$\pm$ 0.0231 \\
B\&C $\mathcal{P}_{\textrm{similarity}}$ & 0.0307$\pm$0.0039
                                                          &0.011$\pm$ 0.0003
                                          						   	& 0.2190$\pm$ 0.0046
                                                           & 0.2147$\pm$0.0212
                                                           & 0.1045$\pm$ 0.0031\\
Blurred Observations $\mathcal{P}_{\textrm{similarity}}$        &   0.1672$\pm$0.0192
                                                    &   0.0707$\pm$0.0074
                                    	              &	0.0351$\pm$0.0072
                                                     & 0.1442$\pm$0.0107
                                                      &   0.2014$\pm$0.0645 \\
Rotation $\mathcal{P}_{\textrm{similarity}}$        &  0.0520$\pm$0.0070
                                                    &   0.0275$\pm$0.0016
                                      	           	&	 0.1020$\pm$0.0115
                                                    & 0.0422$\pm$ 0.0033
                                                    &  0.1020$\pm$0.0115   \\
Shifting $\mathcal{P}_{\textrm{similarity}}$       &  0.0492$\pm$0.0046
                                                   & 0.0650$\pm$0.0092
                                                   & 0.2455$\pm$0.0432
                                                   & 0.0945$\pm$0.0032
                                                   &  0.1167$\pm$0.0121\\
DCT Artifacts $\mathcal{P}_{\textrm{similarity}}$    &  0.0240$\pm$0.0037
                                                            &   0.1325$\pm$0.0301
                                                            &	0.2506$\pm$0.0559
                                                            & 0.2250$\pm$0.0202
                                                            & 0.1592$\pm$0.0369 \\
PT $\mathcal{P}_{\textrm{similarity}}$    &   0.0398$\pm$0.0067
                                                             &   0.012$\pm$0.0007
                                                             &	0.0140$\pm$0.0018
                                                             &  0.0422$\pm$0.0016
                                                             &  0.0440$\pm$0.0050\\
\midrule
C\&W Raw Scores & 15.0$\pm$2.549  & 285.0$\pm$25.495
                            &-20.8$\pm$0.189 & 1168.0$\pm$ 140.696 &  4090.0$\pm$347.979 \\
B\&C Raw Scores & 17.0$\pm$1.651  & 45.0$\pm$6.846  &-21.0$\pm$0.000 & 744.0$\pm$76.957 &  3180.0$\pm$711.027 \\
Blurred Observations Raw Scores & 18.0$\pm$3.405  & 190.0$\pm$33.015  & -21.0$\pm$0.000 & 820.0$\pm$72.013 &  3880.0$\pm$329.484 \\
Rotation Raw Scores & 2.0$\pm$1.264  & 190.0$\pm$ 27.203  & -20.6$\pm$0.209 & 873.0$\pm$201.866 &  3150.0$\pm$482.959 \\
Shifting Raw Scores & 13.0$\pm$1.449  & 70.0$\pm$20.248  & -21.0$\pm$0.000 & 988.0$\pm$ 89.057&  3560.0$\pm$ 437.538\\
DCT Artifacts Raw Scores & 17.0$\pm$3.478  & 60.0$\pm$18.439  & -19.4$\pm$0.428  & 2589.0$\pm$389.679 &  3980.0$\pm$593.936 \\
PT Raw Scores & 1.0$\pm$0.948  & 75.0$\pm$12.649  & -20.9$\pm$0.126 & 486.0$\pm$29.127 &  3550.0$\pm$435.028 \\
\midrule
B\&C [$\alpha,\beta$] &   [1.2,40]  &  [0.9,20]
													&[1.7,40]& [2.4,-275] & [2.4,-260] \\
Blurring Kernel Size &   5  &  3   & 3    & 5 & 5\\
Rotation Degree &   1.4  &  1.6   & 3    & 1.8 & 5\\
Shifting [$t_i,t_j$] &  [1,1]   & [0,1]	&	[2,1] &  [1,2]&  [2,2]  \\
PT Norm &  1   &1	&	3 &2&  3  \\
\bottomrule
\end{tabular}
}
\end{table*}

To probe the deep reinforcement learning policy landscape we will utilize policy dependent worst-case high-sensitivity directions (i.e. adversarial perturbations) as described in Definition \ref{def:highsensitivity} and policy-independent directions innate to the MDP as described in Definition \ref{def:natural}. This probing methodology intrinsically juxtaposes adversarial directions and policy-independent directions with respect to their perceptual similarity distance (see Section \ref{perceptualsim}) to the base states and their degree of impact on the policy performance. More importantly, we question the \textit{imperceptibility} of $\ell_p$-norm bounded adversarial directions in terms of perceptual similarity distance, and compare this \textit{imperceptibility} notion to the policy-independent high-sensitivity directions intrinsic to the MDP.
The fact that policy-independent high sensitivity directions innate to the MDP can achieve ultimately similar or higher drop in the expected cumulative rewards within the perceptual similarity distance brings the line of research focusing on adversarial directions into question. More importantly, the fact that policies trained to resist these adversarial directions and claimed to be "certified" robust are essentially less robust than simple vanilla trained deep reinforcement learning policies as demonstrated in Section \ref{advtrainsec} brings the intrinsic trade-off made during training into question.

While it is possible to interpret the outcomes of contrasting worst-case policy dependent high sensitivity directions (i.e. adversarial) and policy-independent high-sensitivity directions as crucially surprising in terms of the security perspective\footnote{In terms of the security perspective the research conducted in the worst-case high-sensitivity directions in deep reinforcement learning relies heavily on a strong adversary assumption. In particular, this assumption refers to an adversary that has access to the policy's perception system, training details of the policy (e.g. algorithm, neural network architecture, training dataset), ability to alter observations in real time, simultaneous modifications to the observation system of the policy with computationally demanding adversarial formulations as described in Section \ref{advpert} and in Section \ref{advrl}}, our goal is to provide an exact fundamental trade-off made by employing both adversarial attacks and training techniques. The fact that worst-case directions are heavily investigated in deep reinforcement learning research without clear cost and trade-off of these design choices essentially might create bias on influencing future research directions.

To probe the deep neural policy manifold via policy-independent high sensitivity directions we focus on intrinsic changes that are as simple as possible in the high dimensional state representation MDPs. We categorize these changes with respect to their frequency spectrum and below we explain precisely how these high sensitivity directions are computed.

\textbf{Low Frequency Policy-Independent High-Sensitivity Directions:} For the low frequency investigation we utilized brightness and contrast change in the state representations. We have kept moving along high-sensitivity direction as simple as possible as a linear transformation of the base state,
\begin{equation}
\label{bright}
\mathnormal{\hat{s}(i,j) =s(i,j) \cdot \alpha +\beta},
\end{equation}
where $\mathnormal{s(i,j)}$ is the $\mathnormal{ij}^{\textrm{th}}$ pixel of state $\mathnormal{s}$, and $\alpha$ and $\beta$ are the linear brightness parameters. The perspective transform of state representations includes a mapping between four different source and destination pixels given that
\[
\hat{s}(i,j) =s \Bigg(\dfrac{\Gamma_{11}s_i +\Gamma_{12}s_j +\Gamma_{13}}{\Gamma_{31}s_i + \Gamma_{32}s_j+\Gamma_{33}}, \dfrac{\Gamma_{21}s_i +\Gamma_{22}s_j+\Gamma_{23}}{\Gamma_{31}s_i + \Gamma_{32}s_j + \Gamma_{33}}\Bigg)
\]

\begin{equation}
\delta_{k}
\begin{bmatrix}
s^{\textrm{dst}_k}_i \\
s^{\textrm{dst}_k}_j \\
1\\
\end{bmatrix}
= \Gamma \cdot
\begin{bmatrix}
s^{\textrm{src}_k}_i\\
s^{\textrm{src}_k}_j\\
1\\
\end{bmatrix}.
 \label{perseq}
\end{equation}
The norm of a perspective transformation is defined as the maximum distance that one of the corners of the square moves under this mapping. Note that the perspective transformation has effects on both high and low frequencies as also portrayed in Section \ref{ft}.

\begin{figure}[t]
\footnotesize
\centering
\stackunder[0pt]{\includegraphics[scale=0.096]{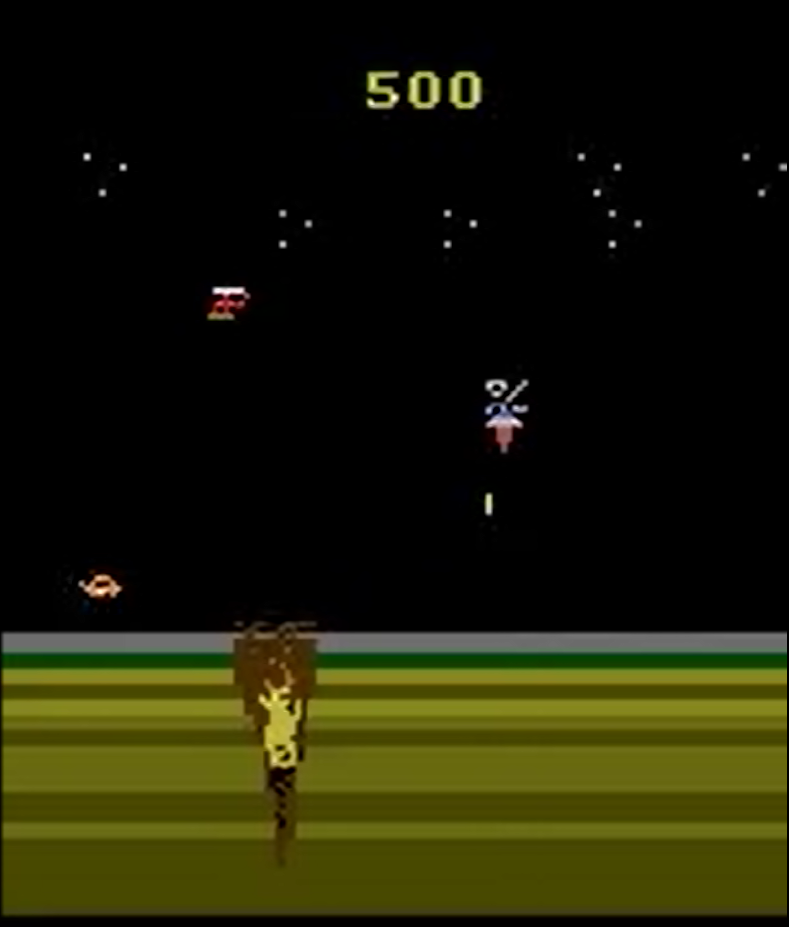}}{}
\hskip 0.1pt
\stackunder[0pt]{\includegraphics[scale=0.048]{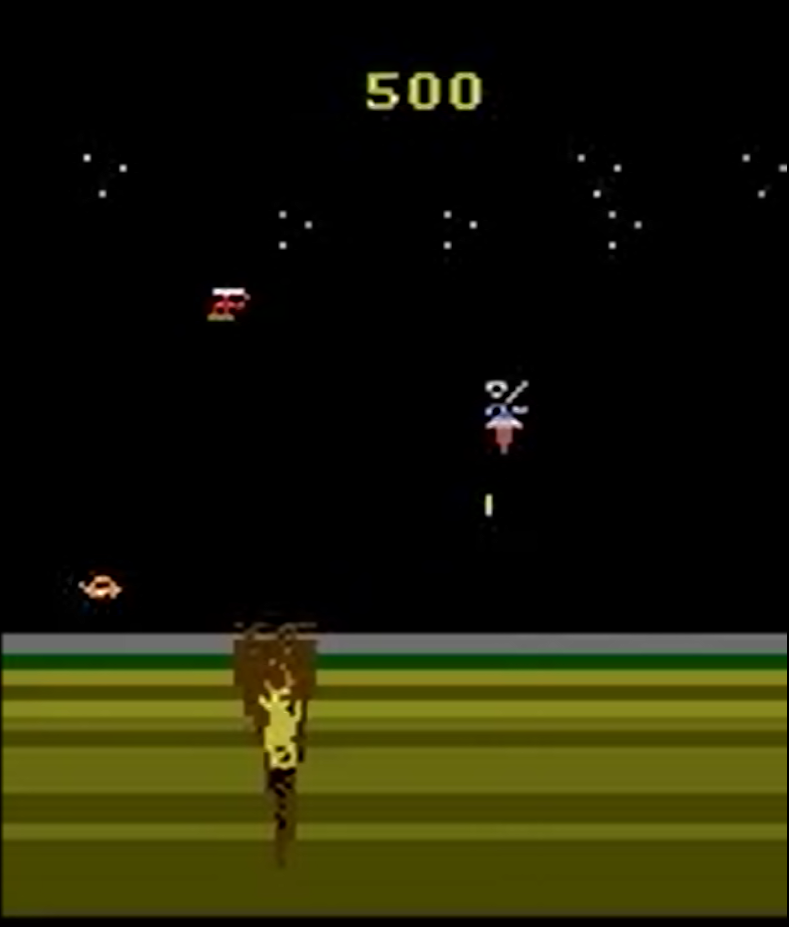}}{}
\stackunder[0pt]{\includegraphics[scale=0.048]{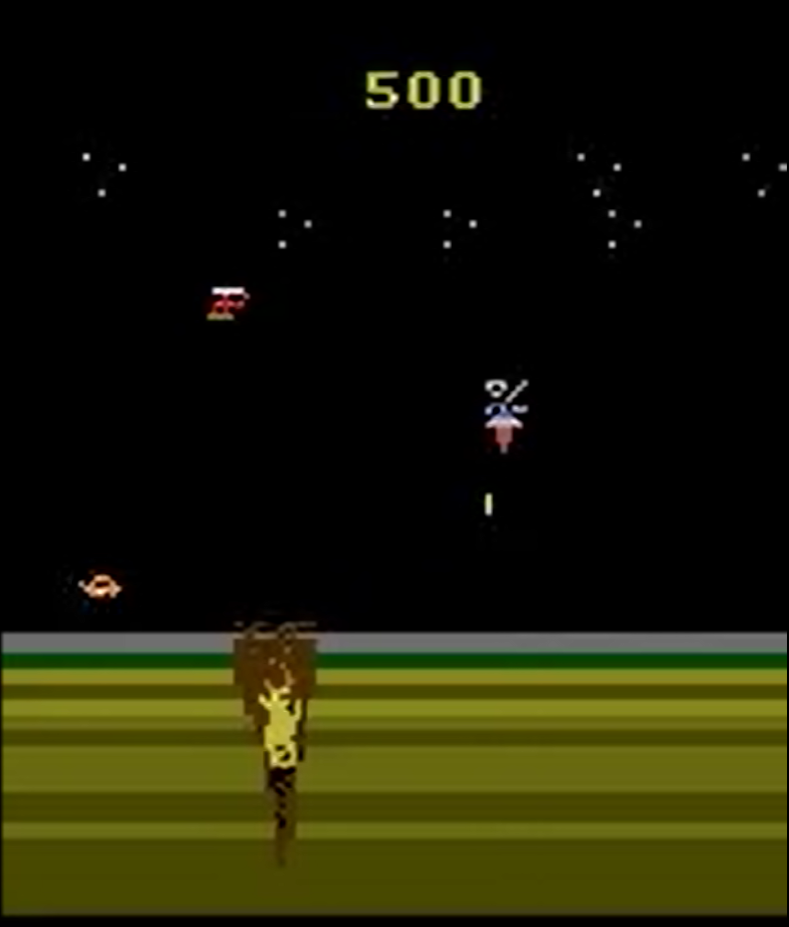}}{}
\stackunder[0pt]{\includegraphics[scale=0.048]{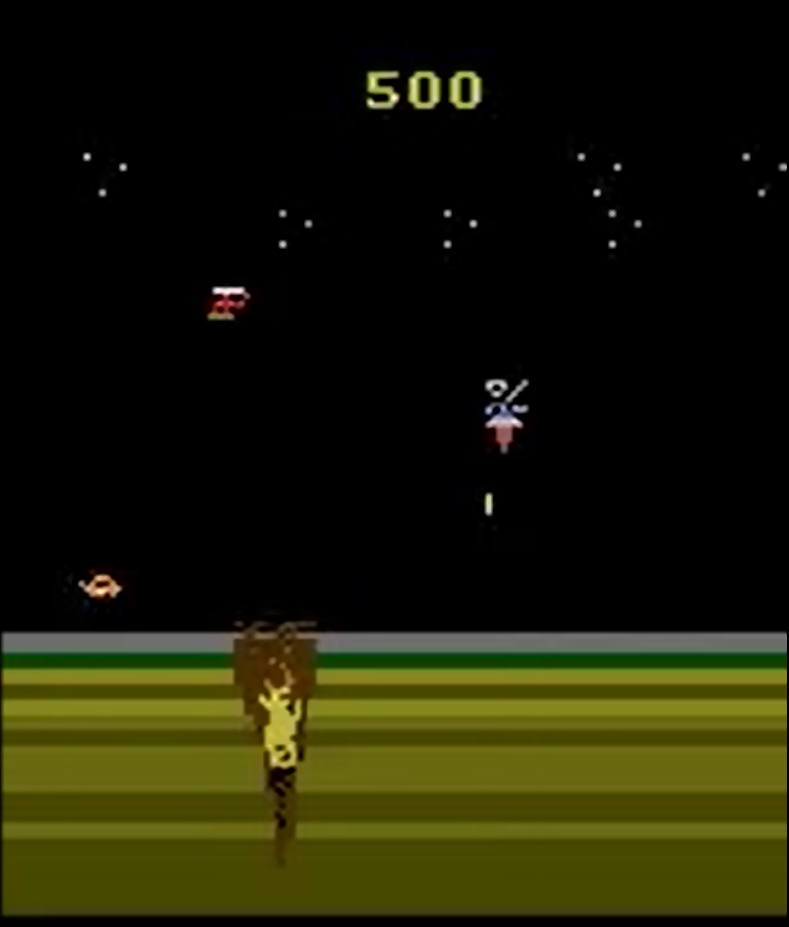}}{}
\stackunder[0pt]{\includegraphics[scale=0.048]{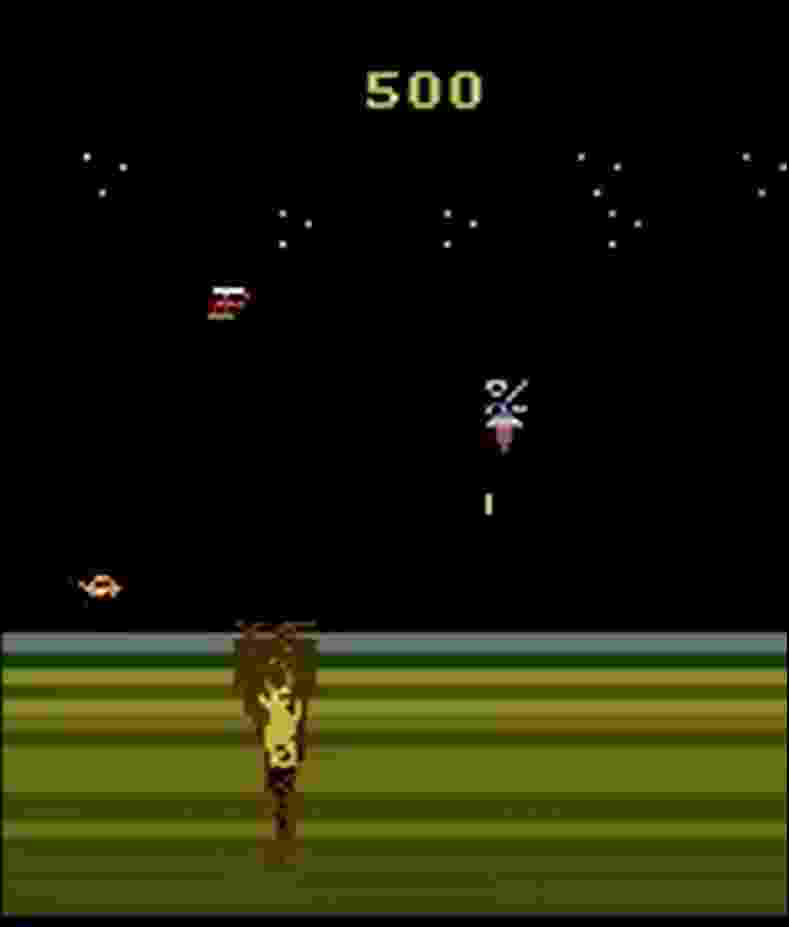}}{}
\stackunder[0pt]{\includegraphics[scale=0.048]{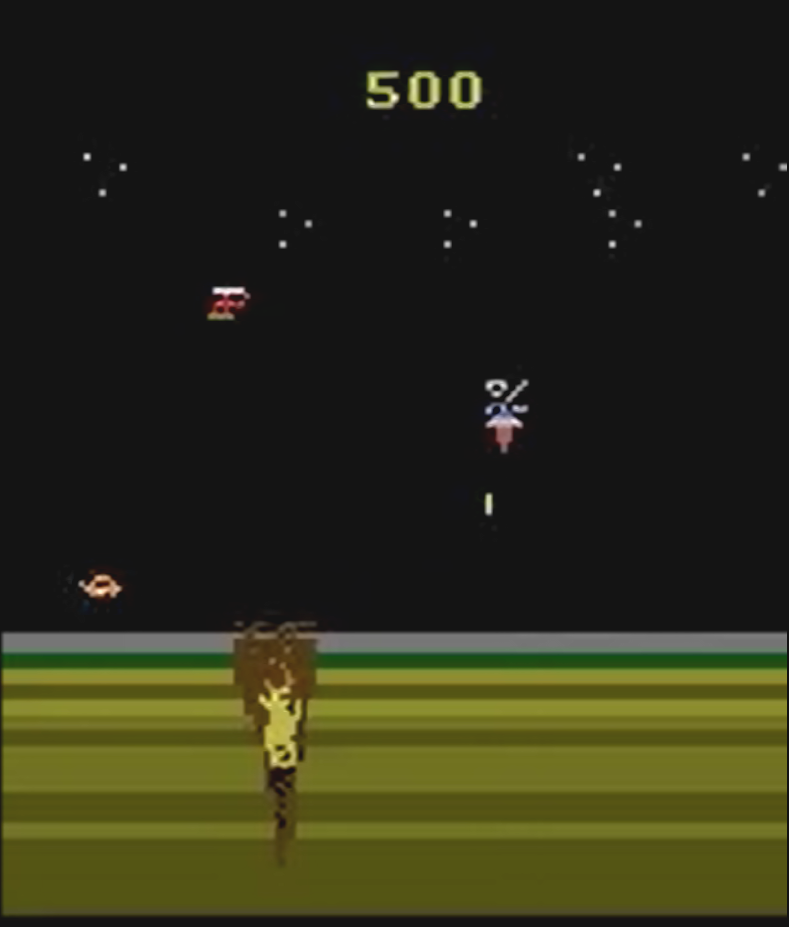}}{} \\
\stackunder[6pt]{\includegraphics[scale=0.1035]{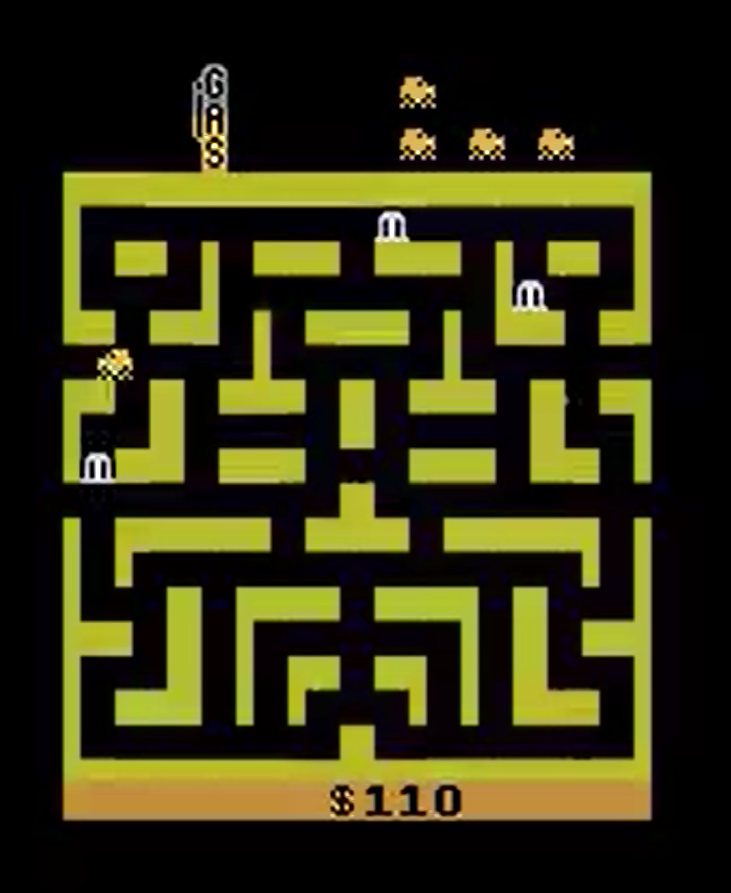}}{\scriptsize{Base State}}
\hskip 0.1pt
\stackunder[6pt]{\includegraphics[scale=0.052]{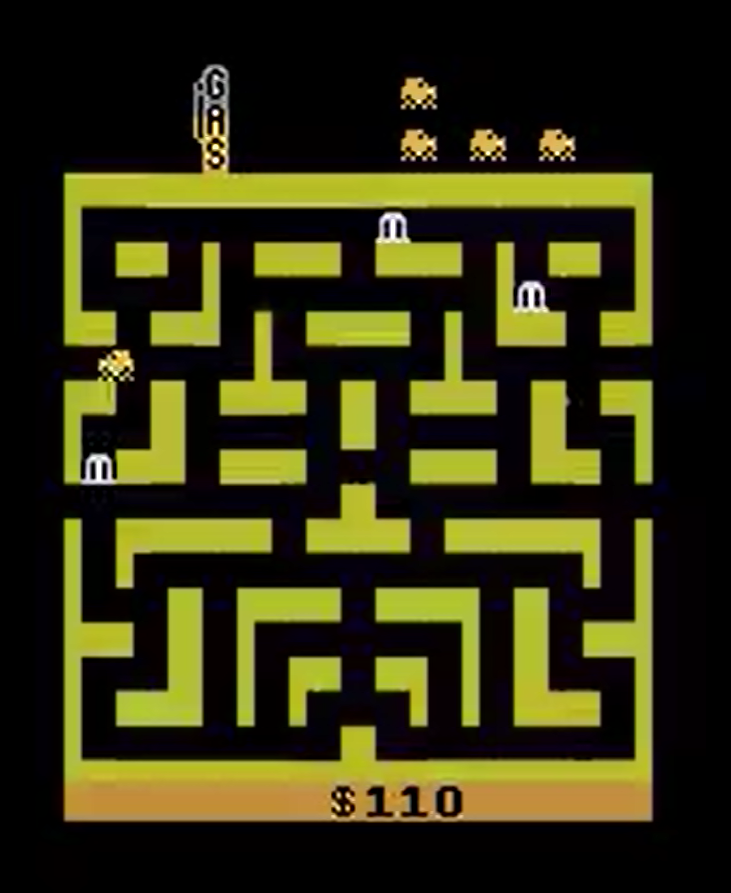}}{\scriptsize{Shift}}
\stackunder[6pt]{\includegraphics[scale=0.052]{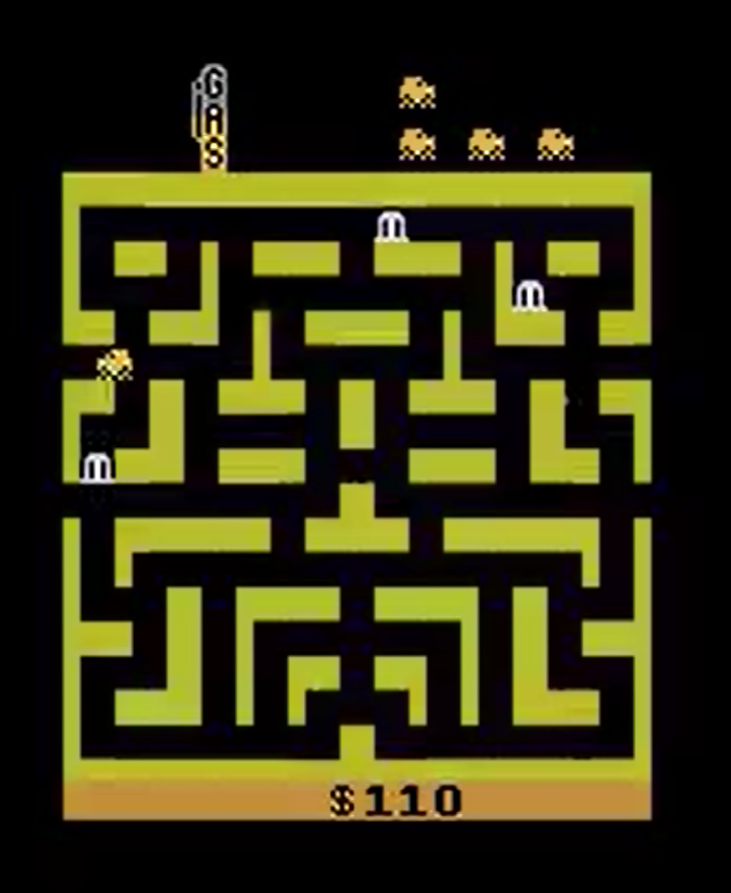}}{\scriptsize{PT}}
\stackunder[6pt]{\includegraphics[scale=0.052]{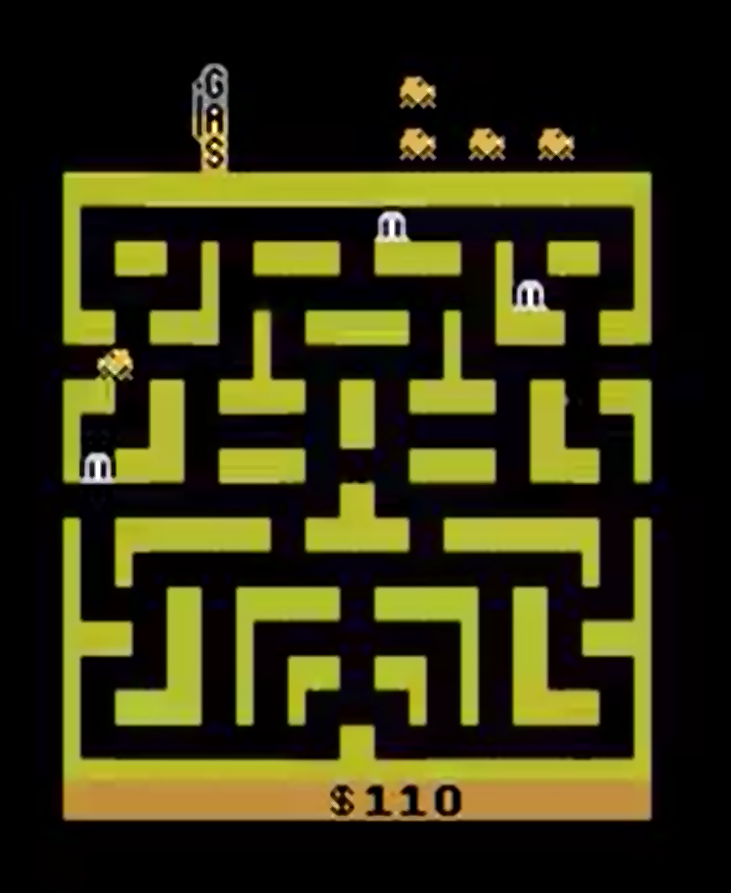}}{\scriptsize{Blur}}
\stackunder[6pt]{\includegraphics[scale=0.052]{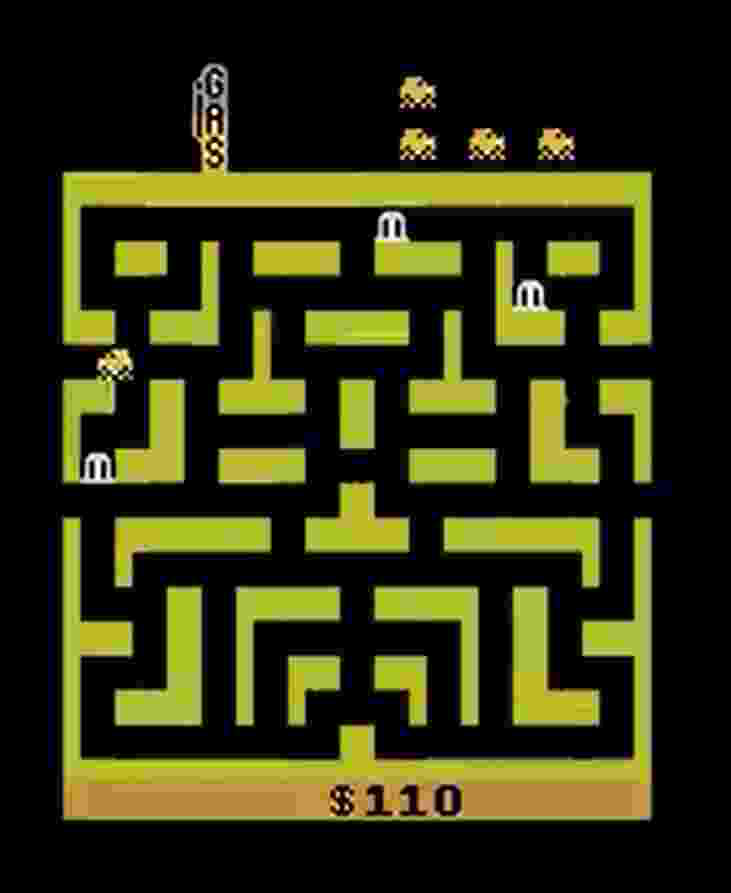}}{\scriptsize{DCT}}
\stackunder[6pt]{\includegraphics[scale=0.052]{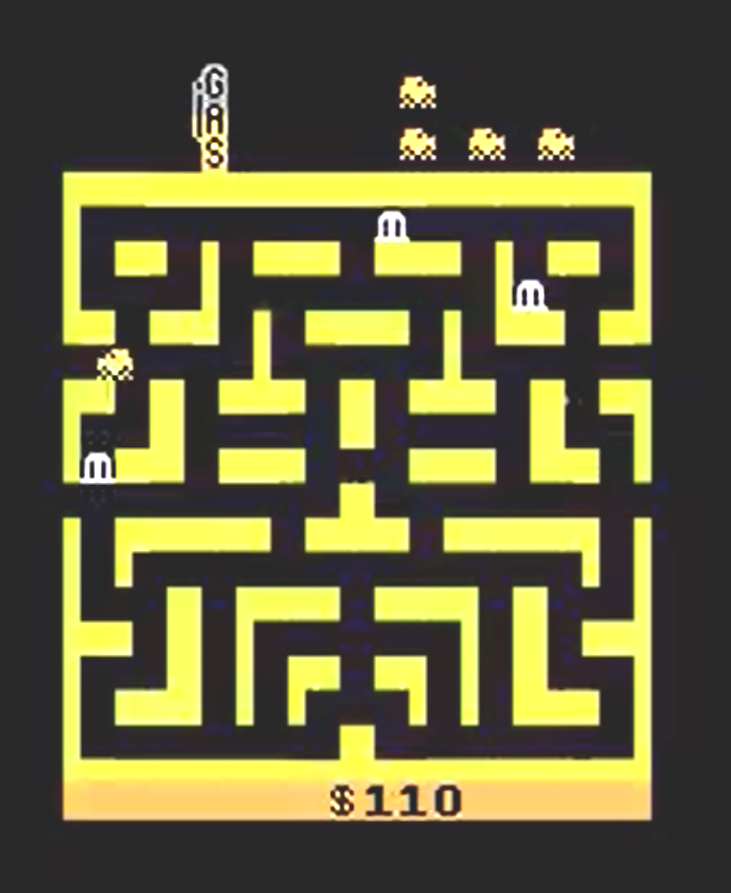}}{\scriptsize{B\&C}} \\
\vskip -0.1in
\caption{Base frame and policy-independent high-sensitivity directions. Columns: base frame, shifting, perspective transformation, blurring, discrete cosine transform artifacts, brightness and contrast. Up: JamesBond. Down: BankHeist. The results for the rest of the MDPs in consideration are reported in the full version of the paper.}
\label{visual}
\vskip -0.1in
\end{figure}

\textbf{High Frequency Policy-Independent High-Sensitivity Directions:} On the high frequency side we included compression artifacts caused by the discrete cosine transform resulting in the loss of high frequency components, also referred to as ringing and blocking artifacts. Another high sensitivity direction considered on the high frequency side of the spectrum is blurring\footnote{Note that in the blurring category one might use several different type of blurring techniques as Gaussian blurring, zoom blurring, defocus blur. Yet all these different types of techniques occupy the same frequency band in the Fourier domain.}. In particular, median blurring which is a nonlinear noise removal technique that replaces the base pixel value with the median pixel value of its neighbouring pixels. In this category kernel size $k$ refers to the fact that the median is computed over a $k \times k$ neighborhood of the base pixel. One of the most fundamental geometric transformations leading to high frequency changes rotates the state observation around the centering pixel with corresponding rotation angle reported as degrees. Lastly, on the geometric transformations, shifting is included, which moves the input in the $x$ or $y$ direction with as few pixels moved as possible. This is denoted with [$t_i,t_j$] as the distance shifted, where $t_i$ is in the direction of $x$ and $t_j$ is in the direction of $y$.

Figure \ref{visual} demonstrates the visual interpretation of moving along these policy-independent high-sensitivity directions innate to the environment described above. While moving along these policy-independent directions is visually imperceptible, we also report exact perceptual similarity distances to the base states computed by Algorithm \ref{alg} in Table \ref{all}.
In more detail, Table \ref{all} shows the raw scores, corresponding performance drops, perceptual similarities to the base states and corresponding hyperparameters for the policy dependent (i.e. adversarial) and policy-independent high-sensitivity sensitivity directions. Hence, the results in Table \ref{all} demonstrate that the policy-independent high-sensitivity directions cause similar or higher degradation in the policy performance within similar perceptual similarity distance. To compute the results in Table \ref{all}, Algorithm \ref{alg} described in Section \ref{main} is utilized.

\section{Moving Along High-Sensitivity Directions in the Adversarially Trained Neural Manifold}
\label{advtrainsec}

In this section we investigate state-of-the-art adversarially trained deep reinforcement learning policies with policy-independent high-sensitivity directions described in Section \ref{main}. In particular, we test State Adversarial Double Deep Q-Network, a state-of-the-art algorithm \citep{huan20}. In this paper the authors propose using what they call a state-adversarial MDP to model adversarial attacks in deep reinforcement learning. Based on this model they develop methods to regularize Double Deep Q-Network policies to be certified robust to adversarial attacks. In more detail, letting $B(s)$ be the $\ell_p$-norm ball of radius $\epsilon$, this regularization is achieved by adding,
\begin{align*}
\mathcal{R}(\theta) = \max\{\max_{\hat{s} \in B(s)}& \max_{a \neq \argmax_{a'} Q(s,a')} Q_{\theta}(\hat{s},a) \\
&- Q_{\theta}(\hat{s},\argmax_{a'} Q(s,a'), -c\}.
\end{align*}
to the temporal difference loss used in standard DQN. In particular, for a sample of the form $(s,a,r,s')$ the loss is
\begin{align*}
  \label{eqn:sadqnupdate}
  \mathcal{L(\theta)} = L_{\mathcal{H}}\left(r + \gamma \max_{a'} Q^{\textrm{target}}(s',a') - Q_{\theta}(s,a)\right) + \mathcal{R}(\theta)
\end{align*}
where $L_\mathcal{H}$ is the Huber loss. Furthermore, we also test the most recent adversarial training technique RADIAL.
In particular, the RADIAL method utilizes interval bound propagation (IBP) to compute upper and lower bounds on the $Q$-function under perturbations of norm $\epsilon$.
In particular, letting $Q^{\textrm{upper}}(s,a,\epsilon)$ and $Q^{\textrm{lower}}(s,a,\epsilon)$ be the respective upper and lower bounds on the $Q$-function when the state $s$ is perturbed by $\ell_p$-norm at most $\epsilon$. For a given state $s$ and action $a$, the RADIAL method utilizes the action-value difference and the overlap given by
\[
Q_{\textrm{diff}}(s,\hat{a}) = \max(0,Q(s,\hat{a}) - Q(s,a)).
\]
The overlap is defined by
\begin{align*}
\mathcal{OV}(s,\hat{a},\epsilon) = \max(0, & Q^{\textrm{upper}}( s,\hat{a},\epsilon) \\
&- Q^{\textrm{lower}}(s,a,\epsilon) + \frac{1}{2}Q_{\textrm{diff}}(s,\hat{a})).
\end{align*}
The adversarial loss used in RADIAL is then given by the expectation over a minibatch of transitions
\[
\mathcal{L}_{\textrm{adv}}(\theta,\epsilon) = \mathbb{E}_{s,a,s'}\left[\sum_{\hat{a}\in A} \mathcal{OV}(s,\hat{a},\epsilon)\cdot Q_{\textrm{diff}}(s,\hat{a})\right].
\]
During training the adversarial loss $\mathcal{L}_{\textrm{adv}}(\theta,\epsilon)$ is added to the standard temporal difference loss. Note that both of these adversarial training algorithms SA-DDQN and RADIAL appeared in NeurIPS 2020 as a spotlight presentation and NeurIPS 2021 consecutively. Thus, it is of great and critical importance in the lines of AI-safety and in terms of affecting overall research progress and effort to outline both the limitations and the actual robustness capabilities of these algorithms.

\begin{table*}[t]
\vskip -0.1in
\caption{The effects of moving along policy-independent high-sensitivity directions in state-of-the-art adversarially trained (SA-DDQN and RADIAL) and vanilla trained deep reinforcement learning policy manifolds.}
\vskip -0.1in
\label{advtrain}
\centering
\scalebox{0.8}{
\begin{tabular}{l|ccc|ccccr}
\toprule
Environment  &
				    &  BankHeist
						&
						&
				    &  Pong
				    &   \\
Training Method &  SA-DDQN
                &  RADIAL
				        &  Vanilla Trained
				        &  SA-DDQN
								&  RADIAL
				        &  Vanilla Trained \\
\midrule
B\&C ($\mathcal{I}$)  & 0.881$\pm$0.010
                      & 0.959$\pm$0.002
                      & 0.971$\pm$0.030
											& 1.0$\pm$0.000
											& 1.0$\pm$0.000
											& 0.996$\pm$0.009    \\
Discrete Cosine Transform Artifacts ($\mathcal{I}$) & 0.960$\pm$0.0014
											                              & 1.0$\pm$0.000
											                              & 0.984$\pm$0.013
																							      & 1.0$\pm$0.000
																										& 1.0$\pm$0.000
																									  & 0.962$\pm$0.032  \\
Perspective Transform ($\mathcal{I}$)  & 1.0$\pm$0.000
                                       &  1.0$\pm$0.000
                                       & 1.0$\pm$0.003
												               & 0.992$\pm$0.0034
																			 & 1.0$\pm$0.000
																			 & 0.996$\pm$0.009  \\
Blurred Observations ($\mathcal{I}$)   & 0.003$\pm$0.002
                                       &  0.985$\pm$0.003
                                       & 0.983$\pm$0.009
								                      	&  0.805$\pm$0.123
																				& 0.901$\pm$0.021
																				&  1.0$\pm$0.000  \\
Rotation ($\mathcal{I}$)                & 1.0$\pm$0.000
																				& 0.992$\pm$0.000
                                        &  1.0$\pm$0.004
								                      	& 1.0$\pm$0.000
																				&  1.0$\pm$0.000
																				&  0.99$\pm$0.015 \\
Shifting ($\mathcal{I}$)               &  1.0$\pm$0.000
																			& 1.0$\pm$0.000
                                       &   0.989$\pm$0.005
									                     & 1.0$\pm$0.000
																			 & 1.0$\pm$0.000
																			 &    1.0$\pm$0.000    \\
\bottomrule
\end{tabular}
}
\vskip -0.1in
\end{table*}

\begin{figure}[h]
\begin{center}
\stackunder[3pt]{\includegraphics[scale=0.14]{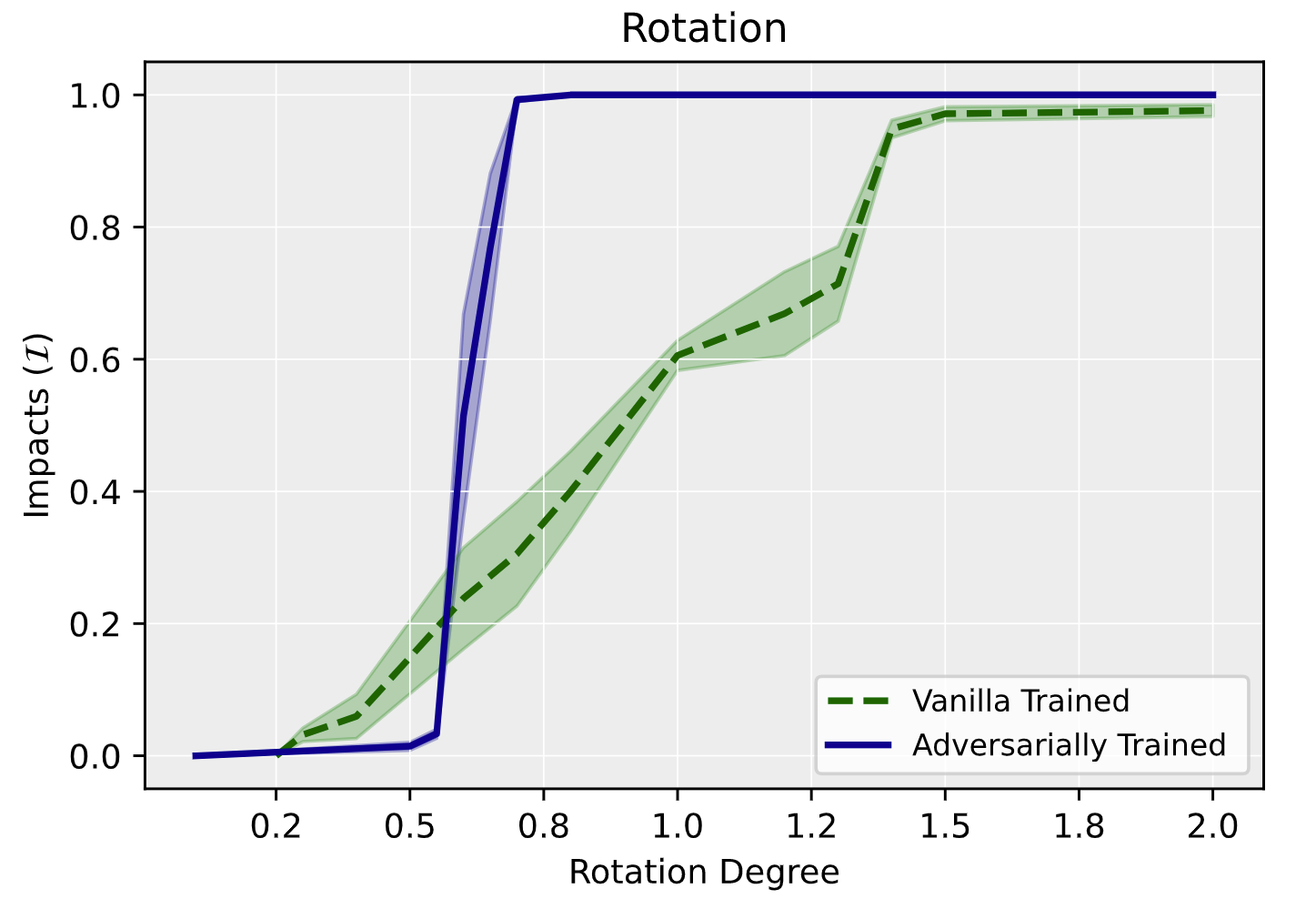}}{}
\stackunder[3pt]{\includegraphics[scale=0.14]{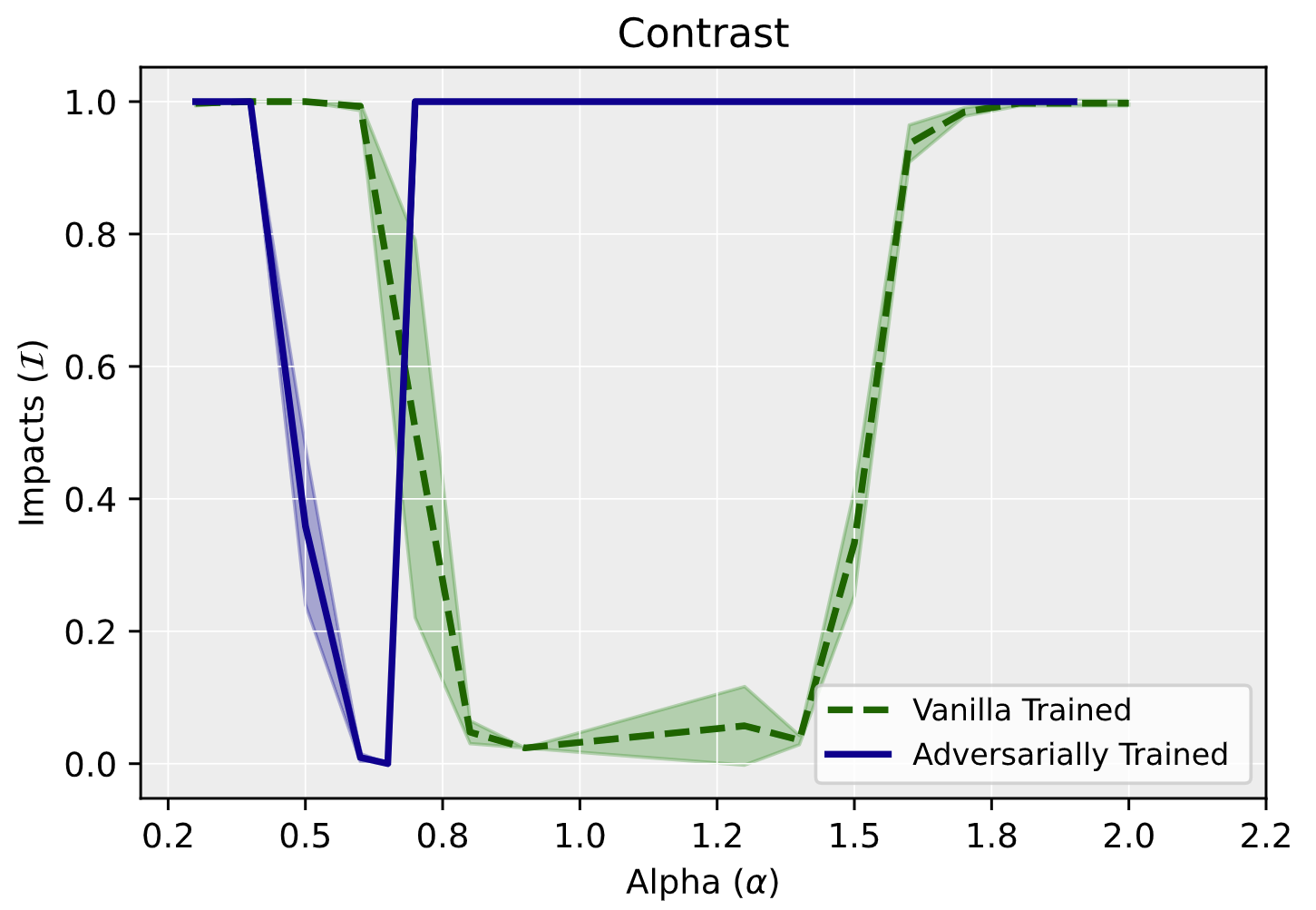}}{} \\
\stackunder[3pt]{\includegraphics[scale=0.15]{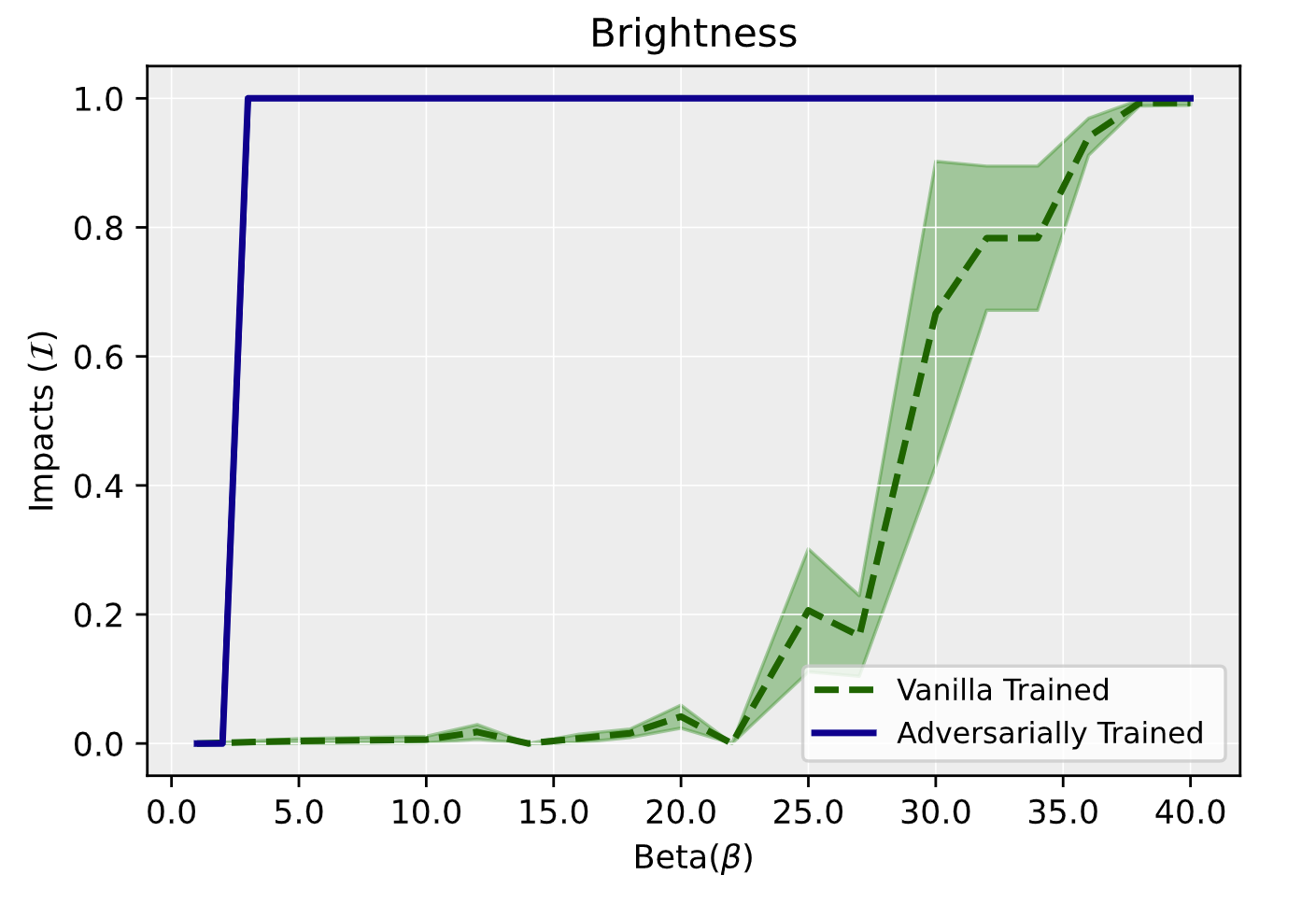}}{}
\stackunder[3pt]{\includegraphics[scale=0.14]{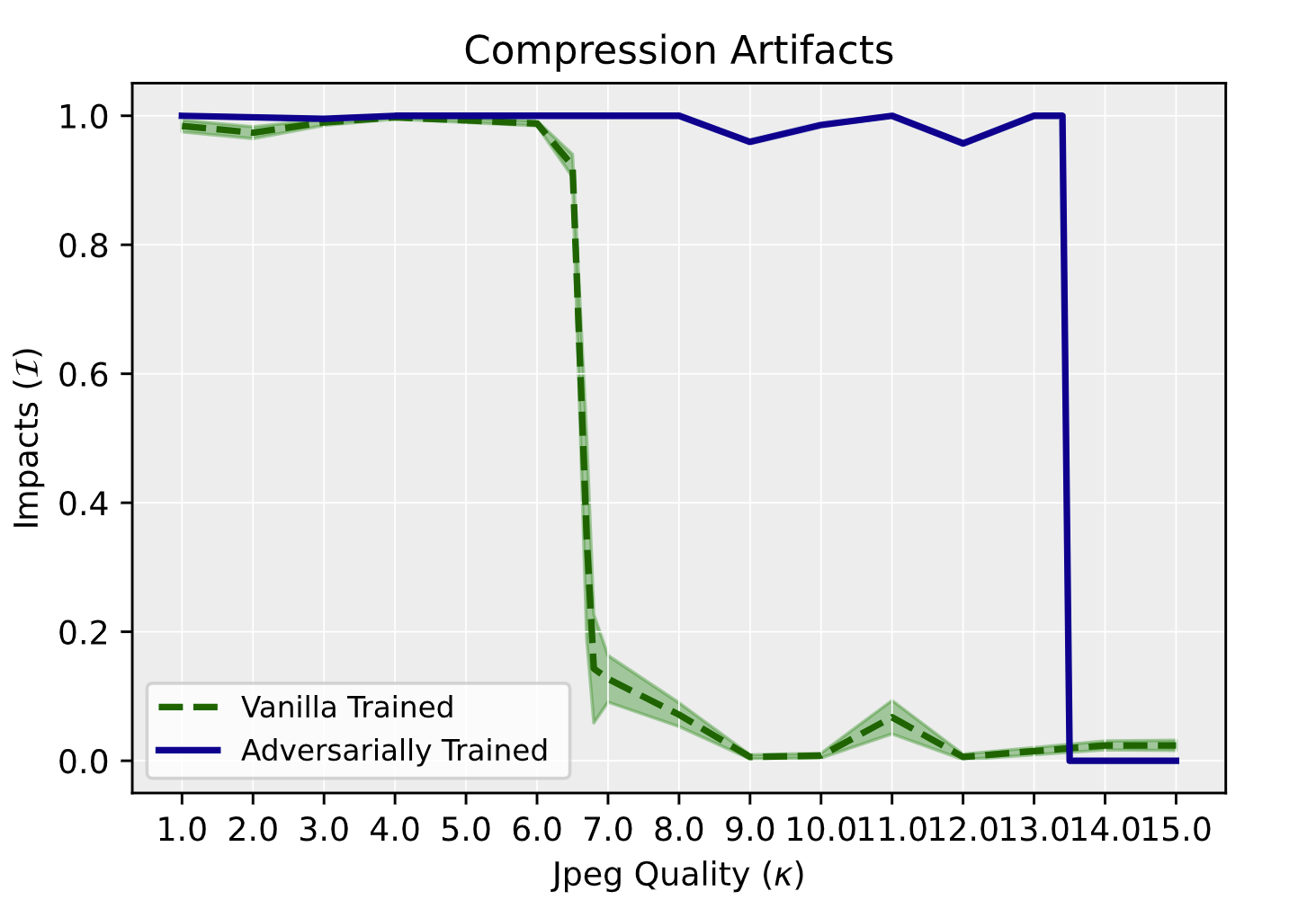}}{}
\end{center}
\vskip -0.2in
\caption{The performance drop results when moved along policy-independent high-sensitivity directions of the state-of-the-art adversarially trained deep reinforcement learning policy manifold and vanilla trained deep reinforcement learning policy manifold with varying the degrees of discrete cosine transform artifacts, brightness, rotation, and contrast.}
\label{advtrainrotation}
\vskip -0.12in
\end{figure}

Table \ref{advtrain} reports the impact values of the policy-independent high-sensitivity directions introduced to the vanilla trained deep reinforcement learning policies and the state-of-the-art adversarially trained deep reinforcement learning policies for both SA-DDQN and RADIAL. Note that the hyperparameters for Table \ref{advtrain} are identical to the hyperparameters in Table \ref{all} for consistency. Thus, the results in Table \ref{advtrain} are not specifically optimized to affect adversarial training. However, Figure \ref{advtrainrotation} reports the effect of varying the amount of movement along  policy-independent non-robust directions, where $\alpha$ stands for contrast, $\beta$ stands for brightness, and $\kappa$ for the level of artifacts caused by the discrete cosine transform. Intriguingly, as these parameters for high-sensitivity directions are varied Figure \ref{advtrainrotation} demonstrates that simple vanilla trained deep reinforcement learning policies are more robust compared to state-of-the-art adversarially trained ones. For instance, modifying brightness with $\beta$ in the range $3.1$ to $20.0$ causes impact close to $1.0$ (i.e. total collapse of the policy) for the adversarially trained policy, but has negligible impact on the vanilla trained policy.

The results in Figure \ref{advtrainrotation} demonstrate that, across a wide range of parameters, adversarially trained neural policies are less robust to natural directions innate to the MDP than vanilla trained policies. This occurs despite the fact that the central purpose of adversarial training is to increase robustness to imperceptible perturbations, where imperceptibility is measured by $\ell_p$-norm. Our results indicate that an increase in robustness to $\ell_p$-norm bounded perturbations can come at the cost of a loss in robustness to other natural types of imperceptible high-sensitivity directions. These results call into question the use of adversarial training for the creation of robust deep reinforcement learning policies, and in particular the use of $\ell_p$-norm bounds as a metric of imperceptibility.

The fact that adversarial training fails to provide robustness has manifold implications. In particular, from the security point of view the effort put into making robust and reliable policies has been misdirected, resulting in policies that are in fact less robust than simple vanilla training. From the alignment perspective, while adversarial training is built to target and make policies safe against adversarial directions, it actually caused these policies to be misaligned with human perception. In terms of foundational understanding of the policies that are being built, our paper brings the term ``robustness'' into question. The decrease in resilience to overall distributional shift that ``certified robust'' adversarial training methods encounter demonstrates the need for further investigation into how robustness should be defined.

\begin{figure}[t!]
\begin{center}
\stackunder[3pt]{\includegraphics[scale=0.064]{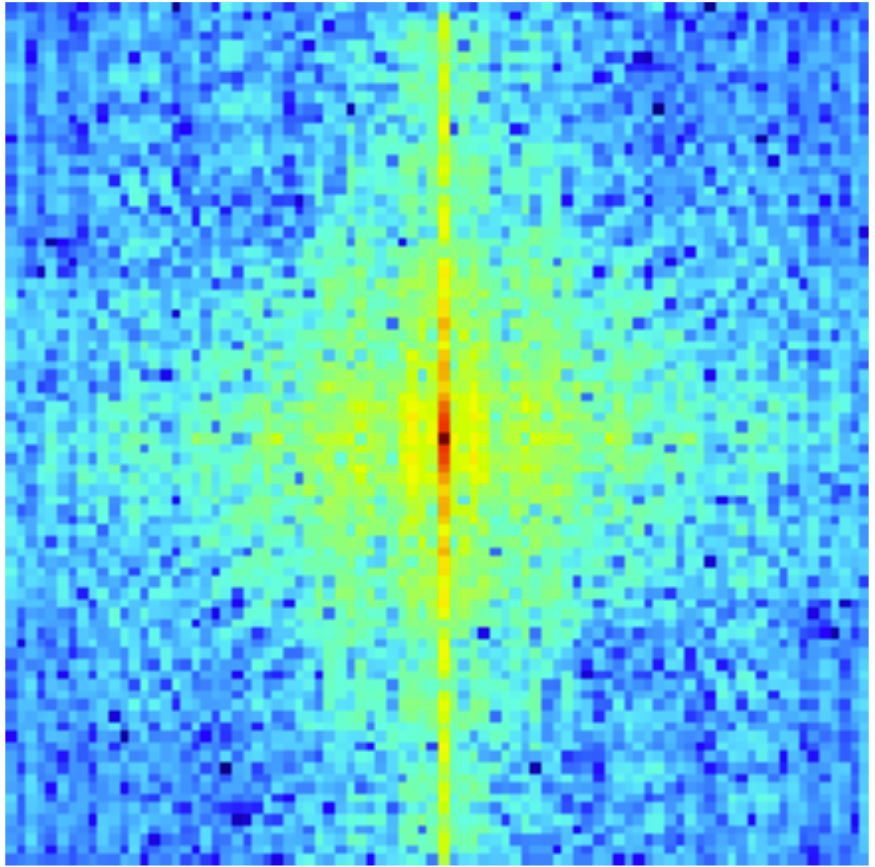}}{}
\stackunder[3pt]{\includegraphics[scale=0.064]{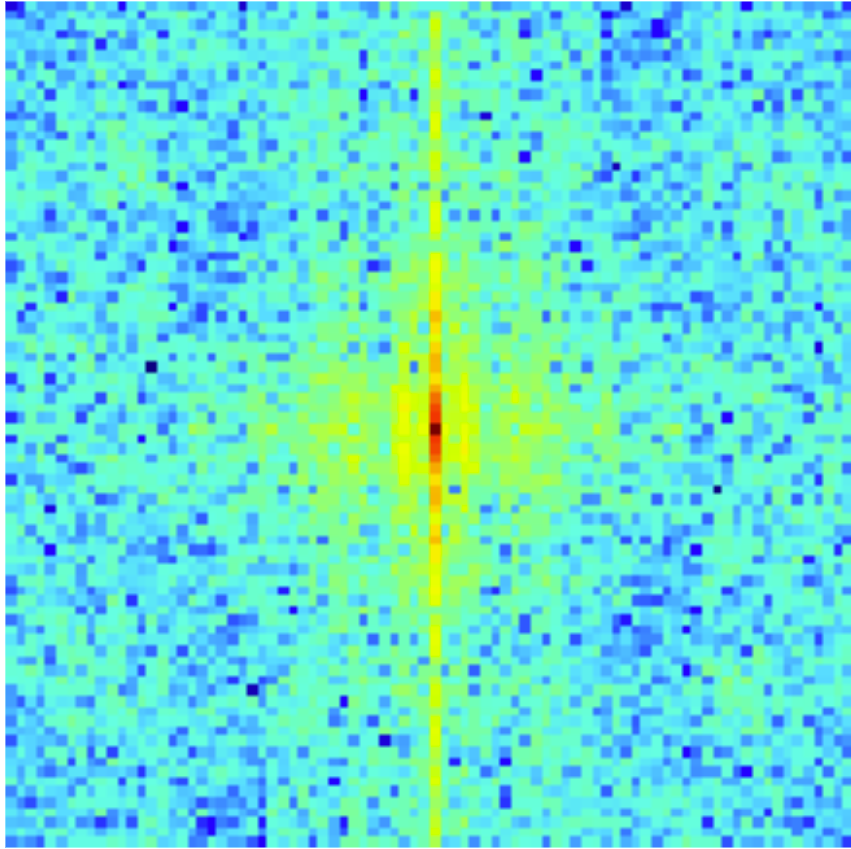}}{}
\stackunder[3pt]{\includegraphics[scale=0.062]{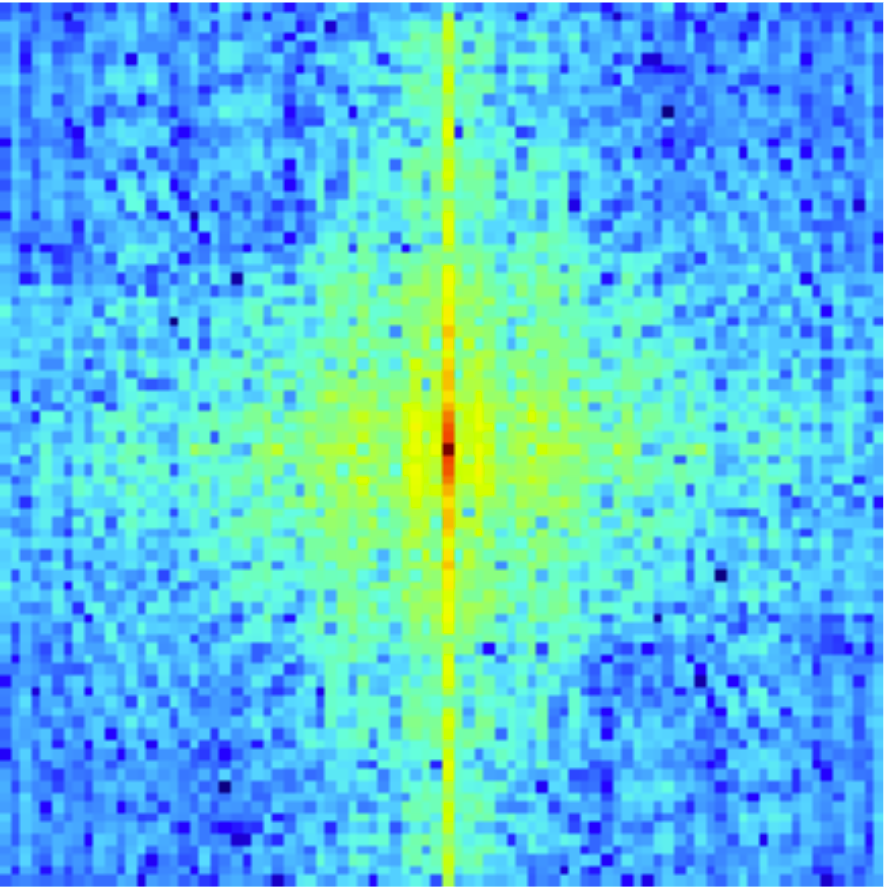}}{}
\stackunder[3pt]{\includegraphics[scale=0.063]{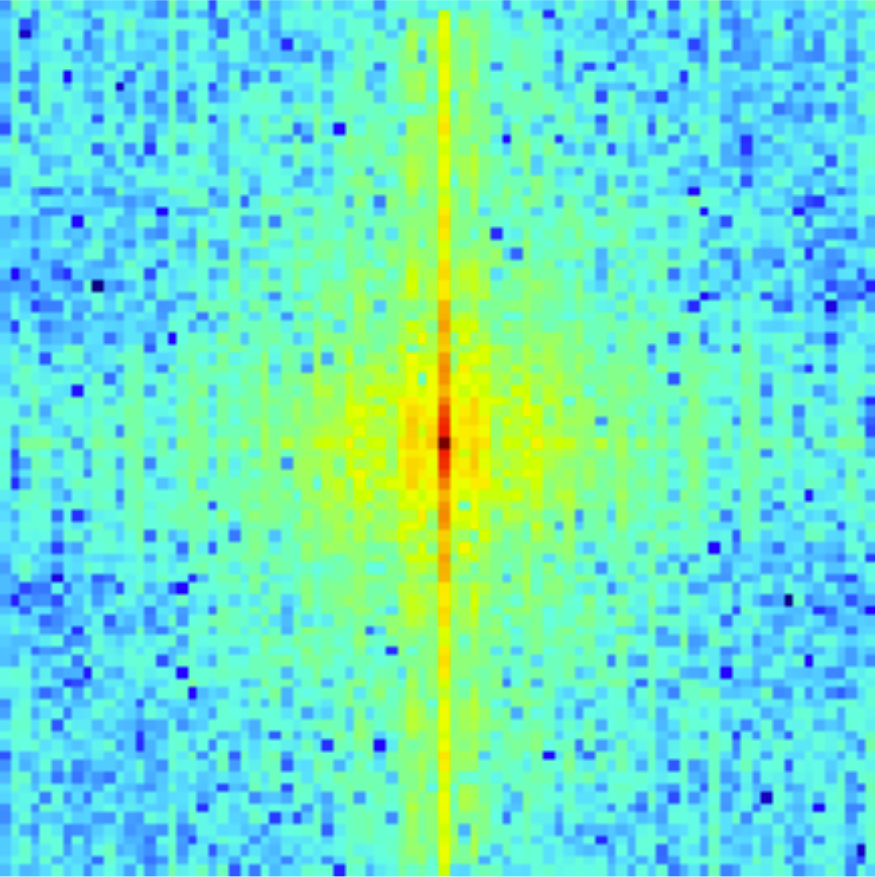}}{}
\stackunder[3pt]{\includegraphics[scale=0.061]{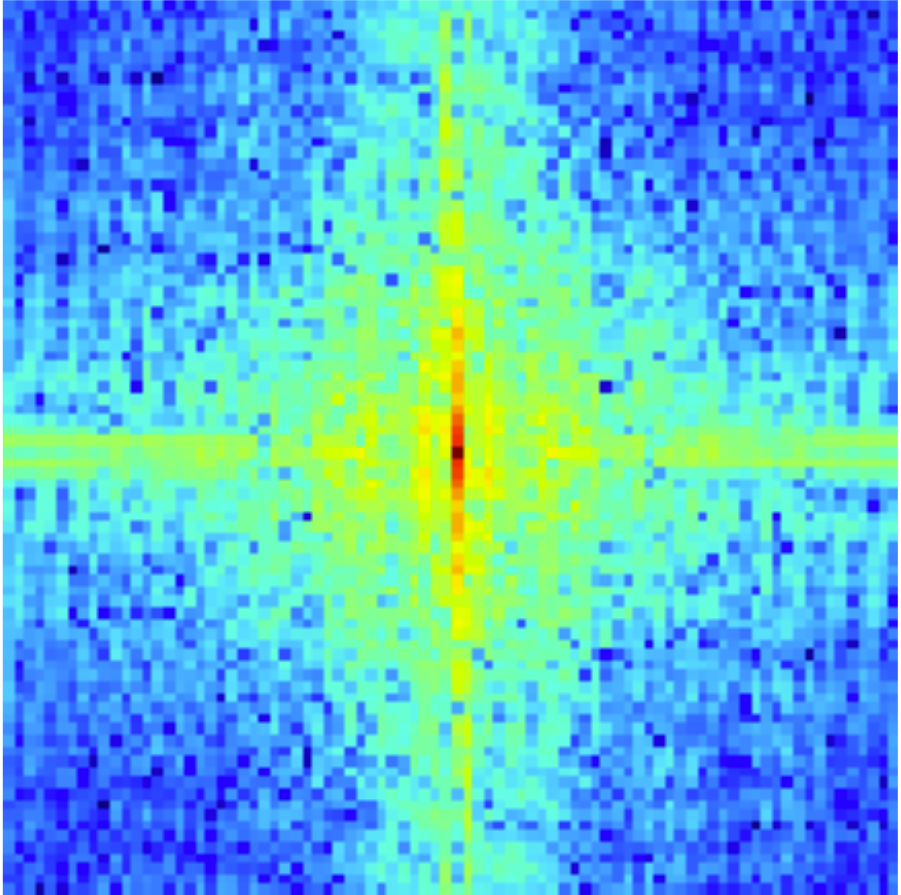}}{}
\stackunder[3pt]{\includegraphics[scale=0.061]{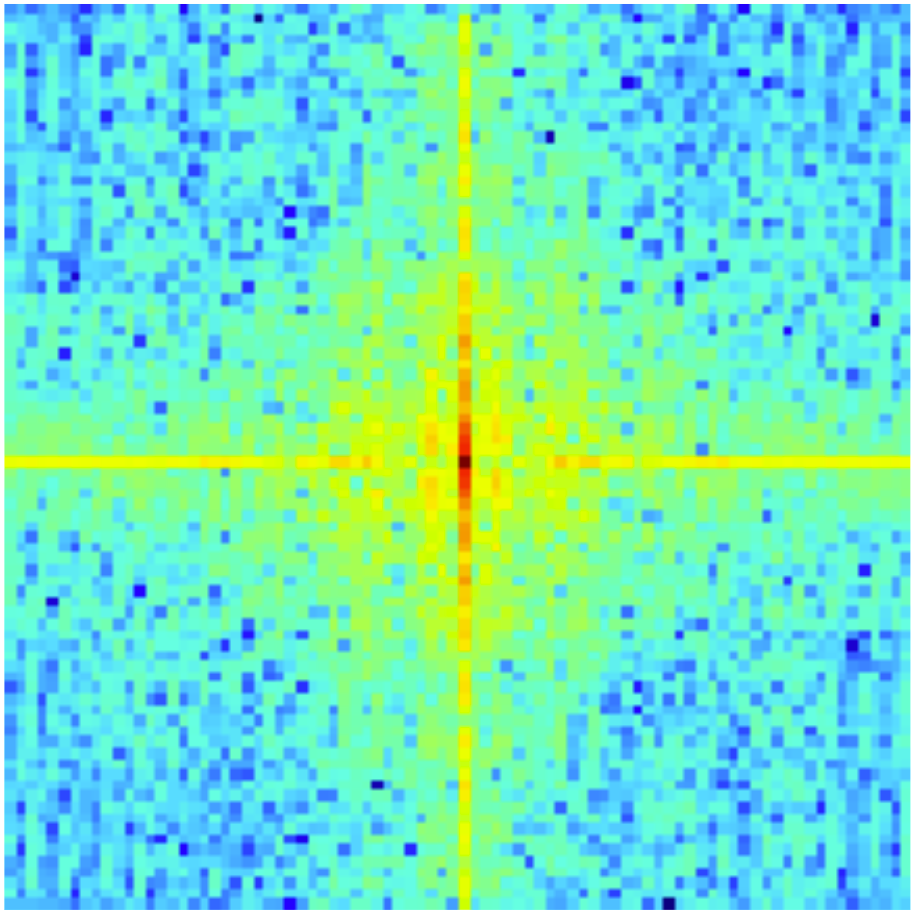}}{}
\stackunder[3pt]{\includegraphics[scale=0.063]{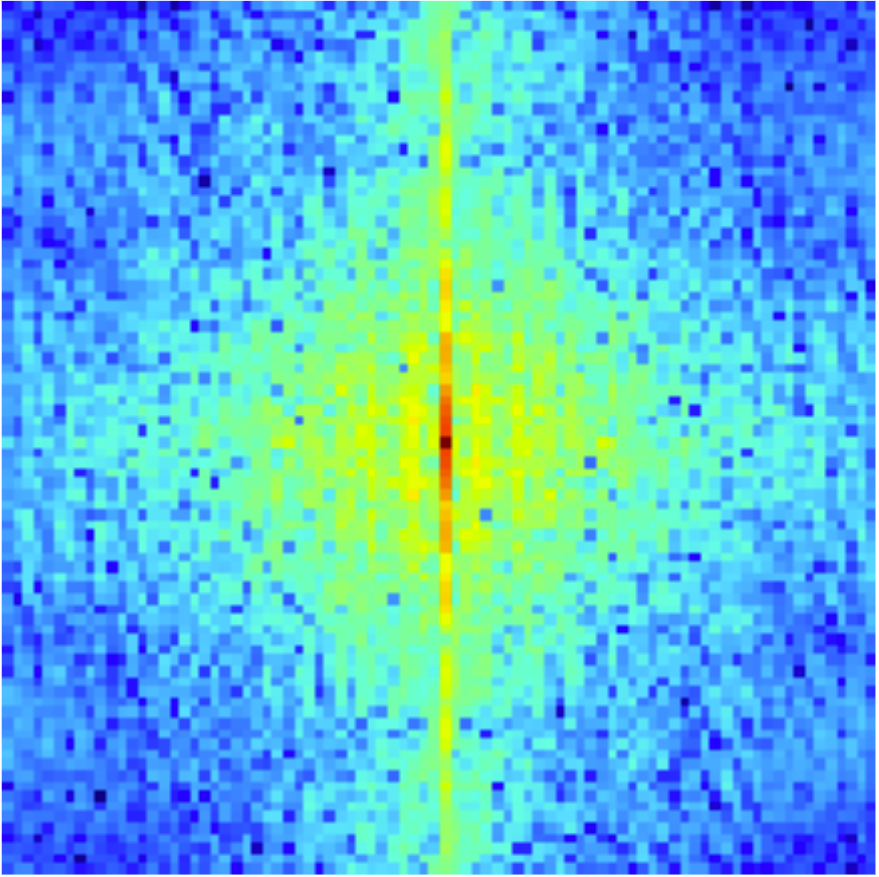}}{}
\stackunder[3pt]{\includegraphics[scale=0.06]{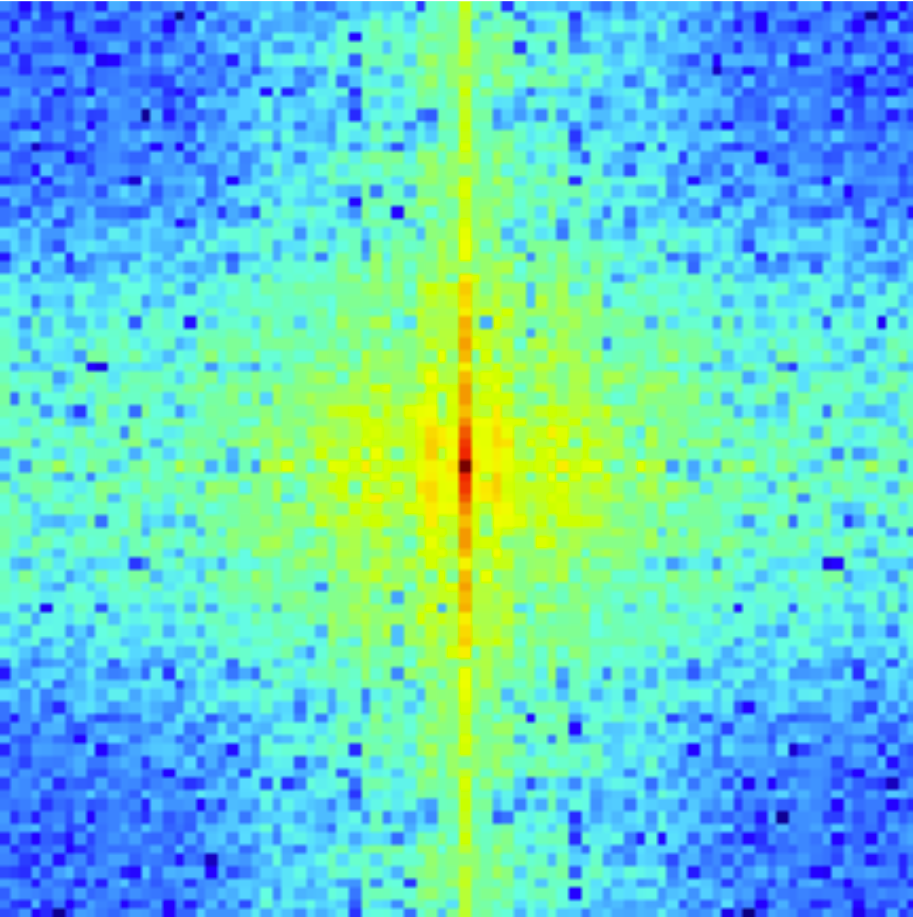}}{}
\end{center}
\begin{center}
\vskip -0.1in
\stackunder[3pt]{\includegraphics[scale=0.0635]{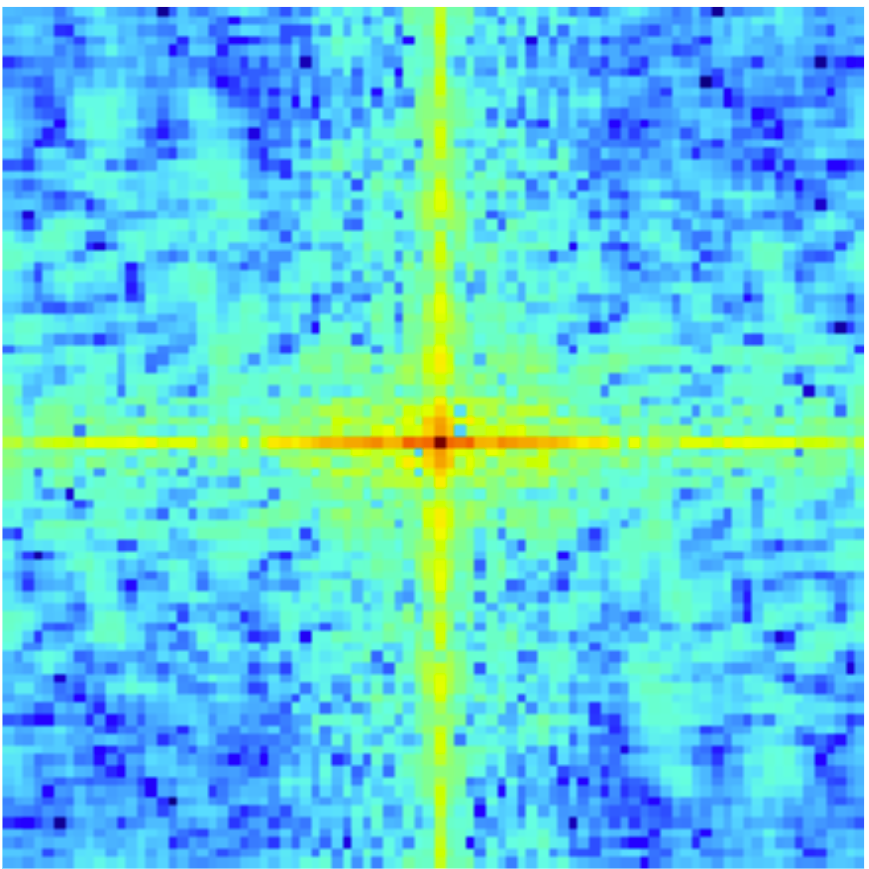}}{\scriptsize{Base}}
\stackunder[3pt]{\includegraphics[scale=0.063]{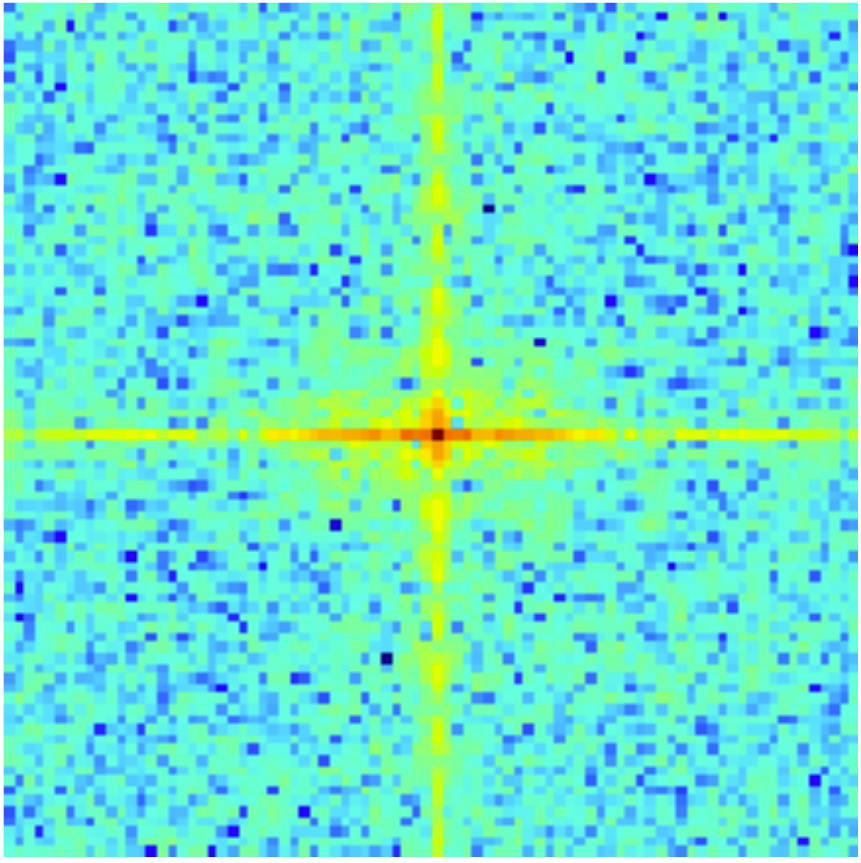}}{\scriptsize{C\&W}}
\stackunder[3pt]{\includegraphics[scale=0.059]{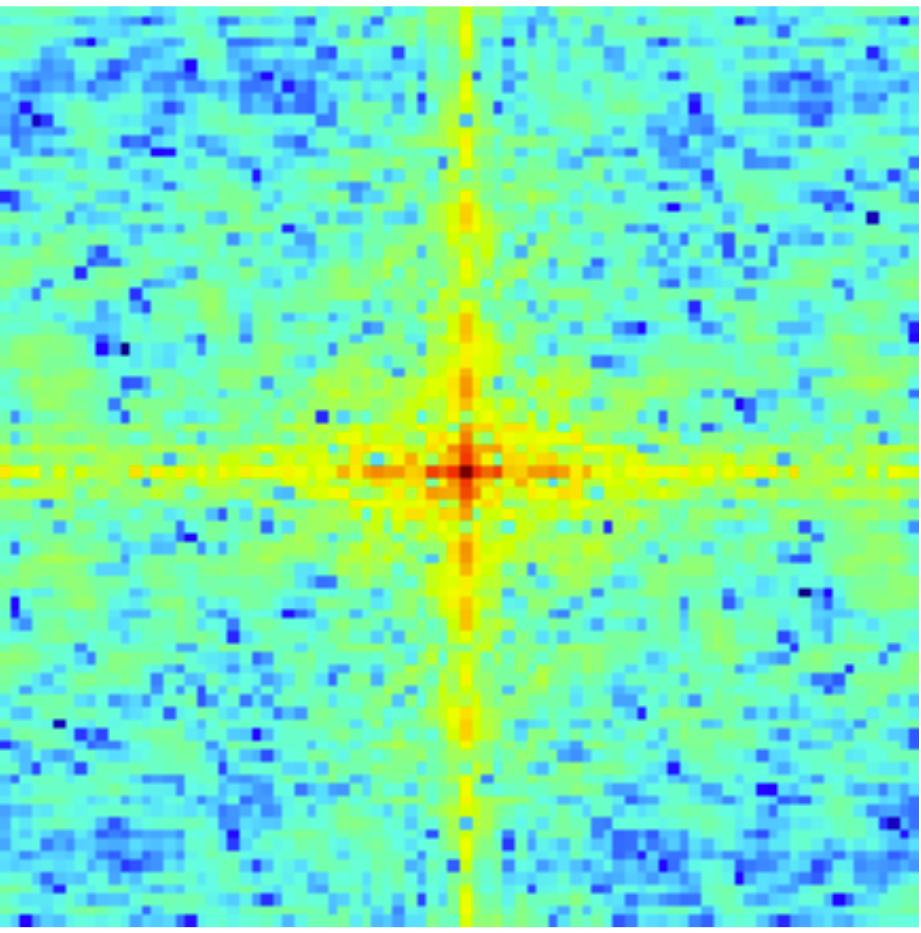}}{\scriptsize{B\&C}}
\stackunder[3pt]{\includegraphics[scale=0.065]{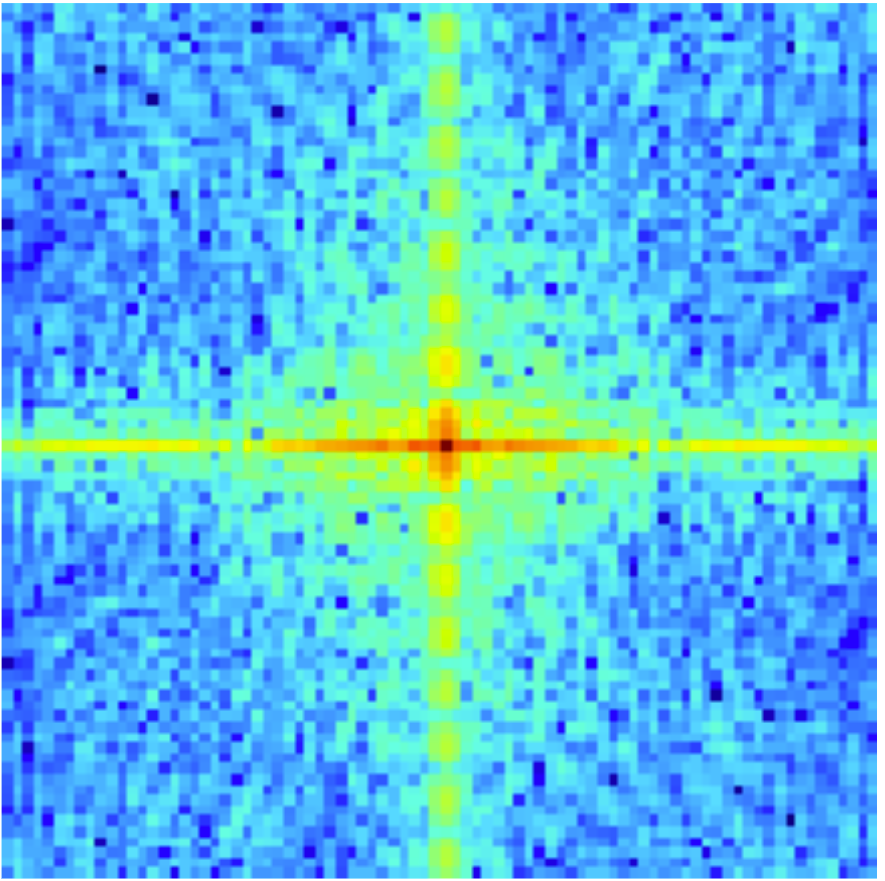}}{\scriptsize{Blur}}
\stackunder[3pt]{\includegraphics[scale=0.063]{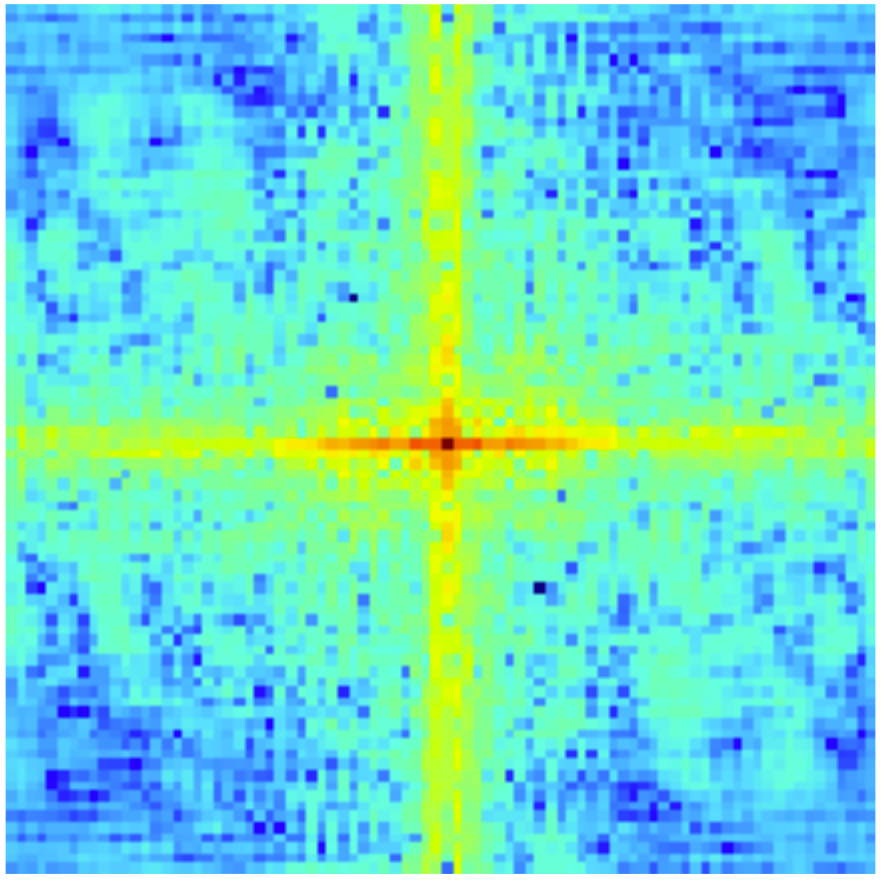}}{\scriptsize{Rotate}}
\stackunder[3pt]{\includegraphics[scale=0.065]{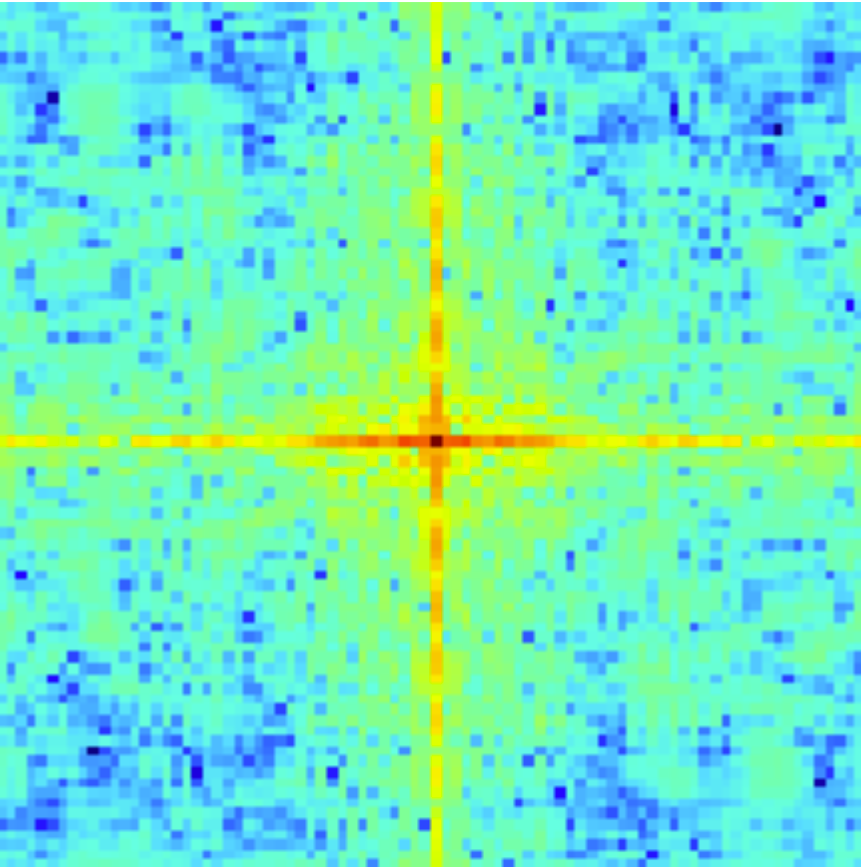}}{\scriptsize{Shift}}
\stackunder[3pt]{\includegraphics[scale=0.069]{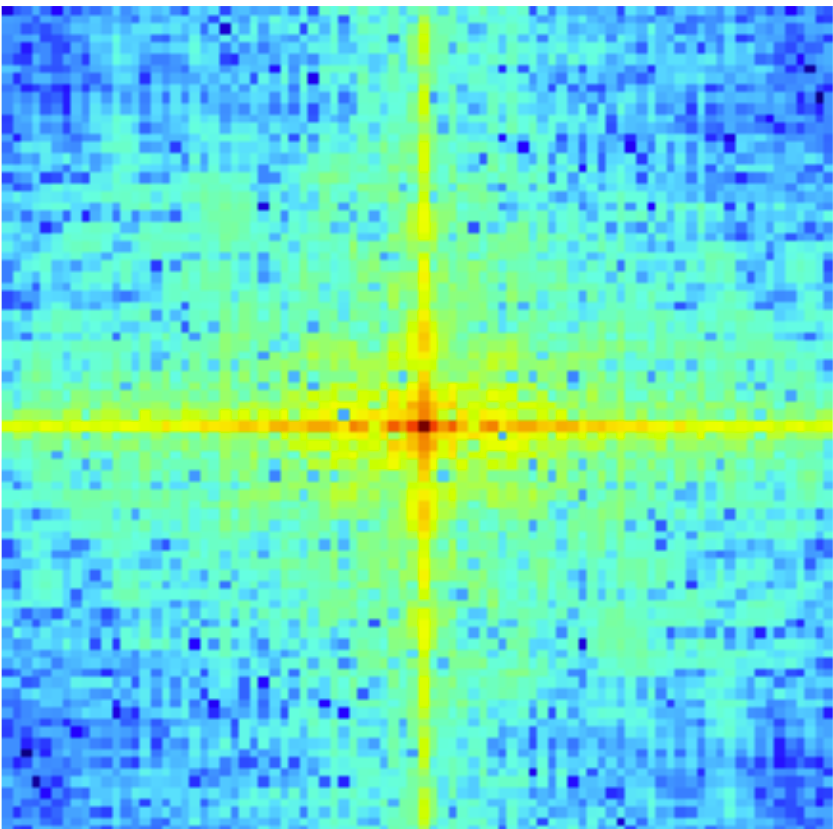}}{\scriptsize{PT}}
\stackunder[3pt]{\includegraphics[scale=0.0615]{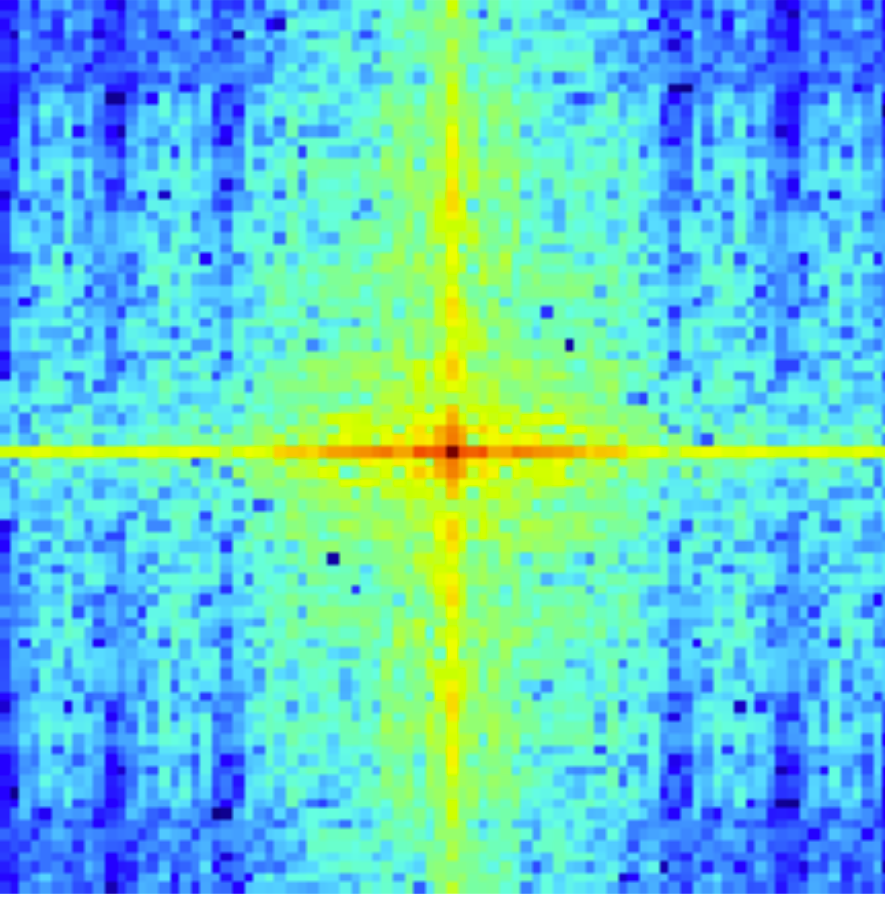}}{\scriptsize{DCT}}
\end{center}
\vskip -0.1in
\caption{Up: $\mathcal{F}_s(u,v)$ for BankHeist. Down: $\mathcal{F}_s(u,v)$ for Riverraid. Columns: base state observation, the Carlini \& Wagner formulation, brightness and contrast, blurred observations, rotation, shifting, perspective transformation, discrete cosine transform artifacts.}
\label{bankfourier}
\vskip -0.1in
\end{figure}

\section{The Frequency Spectrum of the High-sensitivity Directions}
\label{ft}

\begin{figure}[t]
\begin{center}
\stackunder[3pt]{\includegraphics[scale=0.18]{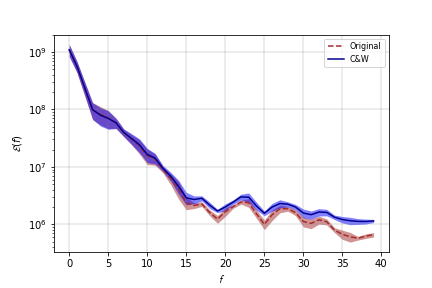}}{}
\stackunder[3pt]{\includegraphics[scale=0.18]{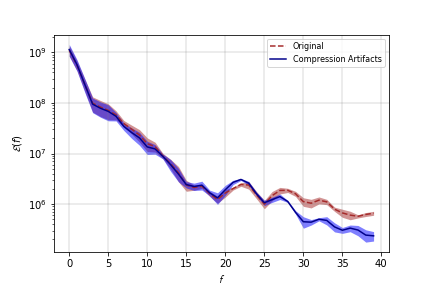}}{}
\stackunder[3pt]{\includegraphics[scale=0.18]{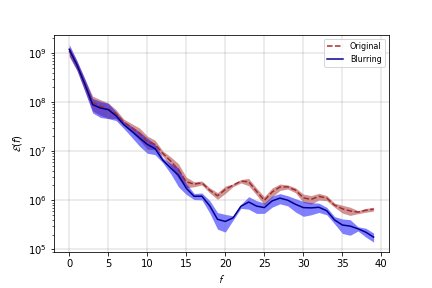}}{} \\
\vskip -0.1in
\stackunder[3pt]{\includegraphics[scale=0.18]{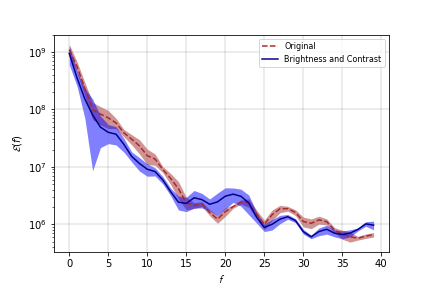}}{}
\stackunder[3pt]{\includegraphics[scale=0.18]{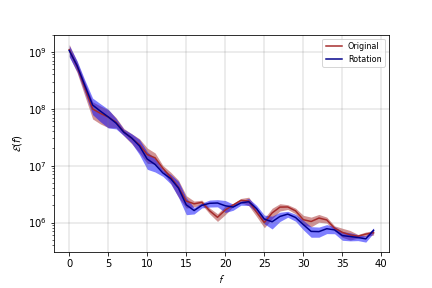}}{}
\stackunder[3pt]{\includegraphics[scale=0.18]{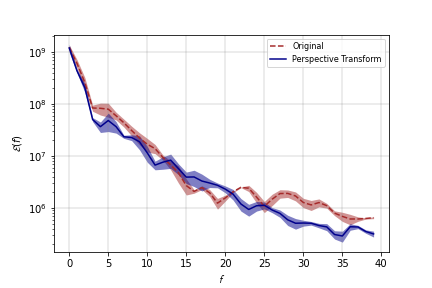}}{}\\
\end{center}
\vskip -0.16in
\caption{Total energy $\mathcal{E}(f)$ spectrum with various perturbations: worst-case directions (C\&W), discrete cosine transform artifacts, perspective transformation, brightness and contrast, shifting, rotation in RiverRaid.}
\label{riverspectral}
\vskip -0.1in
\end{figure}

In this section we provide frequency analysis of the policy dependent worst-case high-sensitivity directions and policy-independent high-sensitivity directions intrinsic to the high dimensional state representation MDP. The purpose of this analysis is to provide quantitative evidence that policy-independent high-sensitivity directions cover a broader portion of the spectrum; thus, provide a broader perspective on robustness than policy dependent adversarial directions alone. In particular, the results in Figure \ref{riverspectral} and \ref{bankfourier} demonstrates how each direction has distinctly different effects in the Fourier spectrum, both policy dependent and policy-independent. In more detail, the frequency spectrum is
\begin{equation}
\mathcal{F}_s(u,v) = \dfrac{1}{IJ} \sum_{i=0}^{I-1} \sum_{j=0}^{J-1} \hat{s}(i,j) e^{-j2\pi(ui/I + vj/J)}
\end{equation}
where $\hat{s}=(s + \xi(s,\pi))$.
Furthermore, we quantify these effects by measuring, for each type of high-sensitivity direction, the change in total Fourier energy at each spatial frequency level.
\begin{equation}
\mathcal{E}(f) = \sum_{\substack{u,v\\ \max\{u,v\}=f}} |\mathcal{F}_s(u,v)|^2
\end{equation}

In Figure \ref{bankfourier} we show the Fourier spectrum of the base state $s$ and the states moved towards high sensitivity directions from the base states $\hat{s}$ with both policy-independent adversarial directions \citep{carlini17}, and the high-sensitivity directions intrinsic to the MDP. In these spectrums the magnitude of the spatial frequencies increases by moving outward from the center, and the center of the image represents the Fourier basis function where spatial frequencies are zero. To investigate which type of high-sensitivity directions occupy which band in the Fourier domain we compute total energy $\mathcal{E}(f)$ for all basis functions whose maximum spatial frequency is $f$. Hence, Figure \ref{riverspectral} shows the power spectral density of the base state compared to states that diverge from base states along the high-sensitivity direction computed via Algorithm \ref{alg} for both policy-independent high-sensitivity directions and policy dependent adversarial directions \citet{carlini17}.

Aside from outlining our methodology, Section \ref{ft} serves the purpose of explaining results obtained in Section \ref{advtrainsec}. In particular, training techniques (e.g. adversarial training) solely focusing on building robustness towards high spatial frequency corruptions become more vulnerable towards corruptions in a different band of the spectrum.
Figure \ref{riverspectral} demonstrates that each policy-independent high-sensitivity direction occupies a different particular band in the frequency domain. In more detail, while the policy dependent adversarial directions increase higher frequencies, the artifacts caused by discrete cosine transform decreases the magnitude of the high frequency band. Along this line both the linear transformation described in \ref{bright} and the geometric transformation described in \ref{perseq} decreases the magnitude of the low frequency band.
The fact that Figure \ref{riverspectral} demonstrates that high-sensitivity directions indeed capture a broader set of directions in the frequency domain assists in providing a wider notion of robustness compared to solely relying on worst-case distributional shifts.

\section{Experimental Details}
\label{experiment}
In our experiments the vanilla trained deep neural policies are trained with Deep Q-Network with Double Q-learning proposed by \cite{hado16} with prioritized experience replay \cite{tom16}, and the adversarially trained deep neural policies are trained via the theoretically justified State-Adversarial MDP modelled State-Adversarial Double Deep Q-Network (SA-DDQN), and with RADIAL (see Section \ref{advtrainsec}) with prioritized experience replay \cite{tom16} with the OpenAI Gym wrapper version \cite{openai} of the Arcade Learning Environment \cite{bell13}. Note that all of the experiments are conducted in policies trained with high dimensional state representations.
To be able to compare between different algorithms and different games the performance degradation of the deep reinforcement learning policy is defined as the normalized impact of an adversary on the agent:
\begin{equation}
\mathcal{I} = \dfrac{\textrm{Score}_{\textrm{clean}}-\textrm{Score}_{\textrm{adv}}}{\textrm{Score}_{\textrm{clean}}-\textrm{Score}^{\textrm{fixed}}_{\textrm{min}}}.
\end{equation}
$\textrm{Score}^{\textrm{fixed}}_{\textrm{min}}$ is a fixed minimum score for a game,  $\textrm{Score}_{\textrm{adv}}$ and $\textrm{Score}_{\textrm{clean}}$ are the scores of the agent  with and without any modification to the agent's observations system respectively. All of the results reported in the paper are from 10 independent runs. In all of our tables and figures we include the means and the standard error of the mean values. More results on the issues discussed in Section \ref{ft} are provided in the full version of the paper with additional high-sensitivity analysis of policy gradient techniques, visualizations of the base states and moving along the high-sensitivity directions intrinsic to the MDP.

\section{Conclusion}

In this paper we focused on probing the deep neural policy decision boundary via both policy dependent specifically optimized worst-case high-sensitivity directions and policy-independent high-sensitivity directions innate to the high dimensional state representation MDPs. We compared these worst-case adversarial directions computed via the-state-of-the art techniques with policy-independent ingrained directions in the Arcade Learning Environment (ALE).
We questioned the \textit{imperceptibility} notion of the $\ell_p$-norm bounded adversarial directions, and demonstrated that the states with minimal ingrained high-sensitivity directions are more perceptually similar to the base states compared to adversarial directions. Furthermore, we demonstrated that the fact that the policy-independent high-sensitivity directions achieve higher impact on policy performance with lower perceptual similarity distance without having access to the policy training details, real time access to the policy's memory and perception system, and computationally demanding adversarial formulations to compute simultaneous perturbations is evidence that high-sensitivity directions are naturally abundant in the deep reinforcement learning policy manifold. Most importantly, we show that state-of-the-art methods proposed to solve robustness problems in deep reinforcement learning are more fragile compared to vanilla trained deep neural policies. We argued for the significance of the interpretations of robustness in terms of the bias it creates in future research directions. Further, while we highlighted the importance of investigating the robustness of trained deep neural policies in a more diverse spectrum, we believe our study can provide a basis for understanding intriguing properties of the deep reinforcement learning decision boundary and can be instrumental in building more robust and generalizable deep neural policies.

\bibliography{example_paper}

\end{document}